\documentclass[10pt]{article} 
\usepackage[accepted]{tmlr}

\usepackage{amsmath,amsfonts,bm}

\def\eqref#1{equation~\ref{#1}}

\def\1{\bm{1}}

\DeclareMathAlphabet{\mathsfit}{\encodingdefault}{\sfdefault}{m}{sl}
\SetMathAlphabet{\mathsfit}{bold}{\encodingdefault}{\sfdefault}{bx}{n}

\DeclareMathOperator*{\argmin}{arg\,min}

\usepackage{hyperref}
\usepackage{url}
\usepackage[capitalize,noabbrev]{cleveref}
\usepackage{subfig}
\usepackage{booktabs}
\usepackage{adjustbox}
\usepackage{pifont}
\usepackage{multirow}
\usepackage{tabularx}
\usepackage{algorithm}
\usepackage{amsmath}
\usepackage{mathtools}
\usepackage{amssymb}
\usepackage{makecell}
\usepackage{tablefootnote}
\usepackage{threeparttable}

\definecolor{mydarkblue}{rgb}{0,0.08,0.45}
\hypersetup{
    colorlinks=true,
    linkcolor=mydarkblue,
    citecolor=mydarkblue,
    filecolor=mydarkblue,
    urlcolor=mydarkblue,
    pdfview=FitH
}

\let\classAND\AND
\let\AND\relax
\usepackage{algorithmic}

\let\AND\classAND
\AtBeginEnvironment{algorithmic}{\let\AND\algoAND}

\newcolumntype{R}[2]{%
    >{\adjustbox{angle=#1,lap=\width-(#2)}\bgroup}%
    l%
    <{\egroup}%
}
\newcommand*\rot{\multicolumn{1}{R{45}{1em}}}
\newcolumntype{P}[2]{%
    >{\adjustbox{angle=#1,lap=\width-(#2)}\bgroup}%
    c%
    <{\egroup}%
}
\newcommand*\rotp{\multicolumn{1}{P{45}{1em}}}
\newcommand{\cmark}{\textcolor{green}{\ding{51}}}
\newcommand{\xmark}{\textcolor{red}{\ding{55}}}
\def\*#1{\mathbf{#1}}

\title{RefinedFields: Radiance Fields Refinement \\ for Planar Scene Representations}

\author{\name Karim Kassab \email k.kassab@criteo.com \\
    \addr Criteo AI Lab, Paris, France \\
    LASTIG, Université Gustave Eiffel, IGN-ENSG, F-94160 Saint-Mandé
    \AND
    \name Antoine Schnepf \email a.schnepf@criteo.com \\
    \addr Criteo AI Lab, Paris, France \\
    Université Côte d’Azur, CNRS, I3S, France
    \AND
    \name Jean-Yves Franceschi \email jycja.franceschi@criteo.com \\
    \addr Criteo AI Lab, Paris, France
    \AND
    \name Laurent Caraffa \email laurent.caraffa@ign.fr \\
    \addr LASTIG, Université Gustave Eiffel, IGN-ENSG, F-94160 Saint-Mandé
    \AND
    \name Jeremie Mary \email j.mary@criteo.com \\
    \addr Criteo AI Lab, Paris, France
    \AND
    \name Valerie Gouet-Brunet \email valerie.gouet@ign.fr \\
    \addr LASTIG, Université Gustave Eiffel, IGN-ENSG, F-94160 Saint-Mandé
}

\def\month{05}  
\def\year{2025} 
\def\openreview{\url{https://openreview.net/forum?id=S6JpSsYBDZ}}

\begin{document}
\maketitle

\begin{abstract}
Planar scene representations have recently witnessed increased interests for modeling scenes from images, as their lightweight planar structure enables compatibility with image-based models.
Notably, K-Planes have gained particular attention as they extend planar scene representations to support in-the-wild scenes, in addition to object-level scenes.
However, their visual quality has recently lagged behind that of state-of-the-art techniques.
To reduce this gap, we propose RefinedFields, a method that leverages pre-trained networks to refine K-Planes scene representations via optimization guidance using an alternating training procedure.
We carry out extensive experiments and verify the merit of our method on synthetic data and real tourism photo collections.
RefinedFields enhances rendered scenes with richer details and improves upon its base representation on the task of novel view synthesis.
Our project page can be found at \url{https://refinedfields.github.io} .

\end{abstract}

\section{Introduction}
\begin{figure}[h]
    \centering
    \subfloat{%
        \setcounter{subfigure}{0}
        \subfloat[RefinedFields Rendering]{\includegraphics[width=0.5\linewidth]{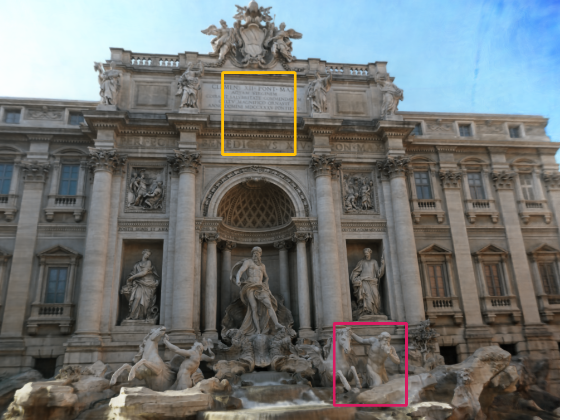}}%
        \begin{tabular}{ccc}
        \vspace{269pt}
        \hspace{-9pt}
        \subfloat{\includegraphics[width=0.165\linewidth]{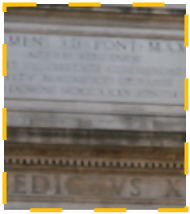}} & \hspace{-15pt}
        \subfloat{\includegraphics[width=0.165\linewidth]{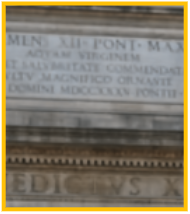}} & \hspace{-15pt}
        \subfloat{\includegraphics[width=0.165\linewidth]{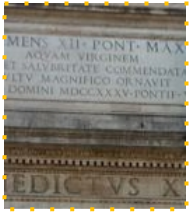}}
        \end{tabular}
    }

    \vspace{-289.5pt}
    \subfloat{%
        \begin{tabular}{ccc}
        \hspace{226pt}
        \setcounter{subfigure}{1}
        \subfloat[K-Planes]{\includegraphics[width=0.165\linewidth]{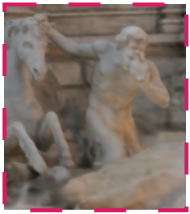}} & \hspace{-15pt}
        \subfloat[RefinedFields]{\includegraphics[width=0.165\linewidth]{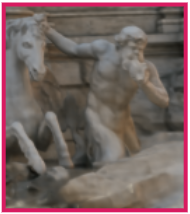}} & \hspace{-15pt}
        \subfloat[GT]{\includegraphics[width=0.165\linewidth]{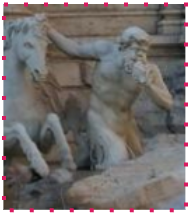}}
        \end{tabular}
    }

    \caption{\textbf{Qualitative Results.} Given images of the Trevi fountain from Phototourism \citep{phototourism}, as well as a pre-trained model \citep{stable-diffusion}, our method leverages the pre-trained model and refines K-Planes with finer details that are under-represented when optimizing the same K-Planes on the images alone.}
    \label{fig:storefront}
\end{figure}

To draw novel objects and views, humans often rely on a blend of \textbf{cognition} and \textbf{intuition}, where the latter is built on a large prior acquired from a long-term continuous exploration of the visual world.
Nevertheless, ablating one of these two elements results in catastrophic representations.
On the one hand, humans find it particularly difficult to draw a bicycle based solely on this preconceived prior \citep{bike}.
However, once one photograph is observed, drawing novel views of a bicycle becomes straightforward.
On the other hand, drawing monuments and complex objects based solely on observed images, and with no preconceived notions of geometry and physics, is also non-trivial. 
In computer vision, recent methods tackling object generation and novel view synthesis generally focus on either the former or the latter ablation.

The first class of methods approaches novel view synthesis by learning scenes through rigorous \textbf{cognition}, as in dense observations of captured images.
Although classic methodologies like structure-from-motion \citep{sfm} and image-based rendering \citep{image-based-rendering} have previously tackled this problem, 
the field has recently seen substantial advancements thanks to stereo reconstruction techniques \citep{dust3r}, as well as implicit neural representation methods \citep{nerf} which this paper focuses on.
Particularly, planar scene representations such as Tri-Planes \citep{eg3d} have recently gained significant attention \citep{3D-nf-gen-triplane-diffusion, LN3Diff} for their lightweight planar structure that allows for a seamless integration with image-based models.
Recent methods extend NeRFs \citep{nerf-w} and Tri-Planes \citep[K-Planes]{kplanes} to support learning from unconstrained ``in-the-wild'' photo collections by enabling robustness against illumination variations and transient occluders.
These representations, however, do not learn any prior across scenes as they are trained from scratch for each scene.
This means that these representations are learned in a closed-world setting, where the information scope is limited to the training set at hand.
Despite the wide interest in planar scene representations, their visual quality currently falls short behind recent state-of-the-art methods \citep{wildgaussians,splatfacto-w}.
Our goal is to integrate pre-trained networks into the training framework of K-Planes, with an aim to improve their rendering performance.

The second class of methods tackles novel view synthesis and object generation by learning and leveraging priors over images and scenes, reminiscent of drawing from insights and \textbf{intuition}.
These methods have also recently witnessed accelerated advancements.
Recent works leverage pre-trained networks to achieve Object Generation (OG) \citep{dreamfusion, latentnerf} and Novel View Synthesis (NVS) \citep{pixelnerf, dietnerf, zero-1-to-3, realfusion}.
Particularly, \citet{zero-1-to-3} achieve NVS from single images only by simply fine-tuning a pre-trained latent diffusion model \citep{stable-diffusion}.
This proves pivotal significance related to large-scale pre-trained vision models, as it shows that, although trained on 2D data, these models learn a rich geometric 3D prior about the visual world.
Nevertheless, as these pre-trained models alone have no explicit multi-view geometric constraints, their use for 3D applications is usually prone to geometric issues (e.g.\  geometric inconsistencies, multi-face Janus problem, content drift issues \citep[Figure~1]{mvdream}).
This class of methods has not yet been explored to enhance implicit models representing in-the-wild scenes, as leveraging priors over such representations is not evident.

\textbf{RefinedFields proposal.}
Our work builds on the previous discourse and aims to enhance planar scene modeling by leveraging pre-trained networks.
To this end, we present an alternating training algorithm that integrates a novel \emph{scene refining} stage, which aims to propose a better optimization initialization by utilizing a pre-trained model.
We adopt K-Planes \citep{kplanes} as a base scene representation, which extend Tri-Planes to support in-the-wild scenes, in addition to object-level scenes.
\citet{pi3d} highlight that features in planar scene representations resemble projected scene images, a finding we also corroborate in \cref{x:inspection}, and exploit to improve upon K-Planes.
Specifically, RefinedFields \emph{refines} planar scene representations by projecting them onto the space of representations inferable by a pre-trained network, which pushes K-Planes features to more closely resemble real-world images. 
To do so, we build on the seamless integration of K-Planes with image-based models and present an alternating training procedure that iteratively switches between optimizing a K-Planes representation on images from a particular dataset, and fine-tuning a pre-trained network to output a new conditioning leading to a refined version of this K-Planes representation. 
Overall, this procedure guides the optimization of a particular scene, by leveraging not only the training dataset at hand but also the rich prior lying within the weights of the pre-trained model, which is a first for in-the-wild scene modeling.

We conduct extensive quantitative and qualitative evaluations of RefinedFields.
We show that our method improves upon K-Planes with richer details in scene renderings.
We prove via ablation studies that this added value indeed comes from the fine-tuned prior of the pre-trained network.
\cref{fig:storefront} illustrates the improvements our method showcases on the \emph{Trevi fountain} scene from Phototourism \citep{phototourism}.

\section{Related Work}
\label{sec:related-work}
RefinedFields achieves geometrically consistent novel view synthesis, that can also be applied in-the-wild, by leveraging K-Planes, and a large-scale pre-trained network.
Our method is the first to satisfy all of these attributes, as summarized in \cref{table:comparison}. 
In this section, we develop the various preceding works from which our method takes inspiration.

\paragraph{Neural representations.}
Neural rendering \citep{sota-nr} has seen significant advancements since the introduction of Neural Radiance Fields \citep[NeRF]{nerf}. 
At its core, NeRF learns a scene by fitting the weights of a neural network on posed images of said scene.
This subsequently enables the reconstruction of the scene thanks to volume rendering \citep{volume-rendering}. 
Subsequent to NeRFs, various scene representations have been introduced \citep{tensorf, instantngp, gaussian-splatting}.
Particularly, \citet{eg3d} introduce Tri-Planes, a planar scene representation serving as a middle ground between implicit and explicit representations, enabling a faster learning of scenes.
Tri-Planes have been widely adopted in recent works \citep{3D-nf-gen-triplane-diffusion, LN3Diff, rodin}, as their planar structure enables a seamless integration with image-based models.

\begin{table}[t]
\centering
\caption{
    \textbf{Related work overview.} RefinedFields leverages a pre-trained prior \citep{stable-diffusion} on the scene representation to refine K-Planes, our underlying scene representation, utilized for novel view synthesis in the wild (NVS-W). \\
    $^{\dagger}$Non exhaustive, other works with characteristics similar to NeRF-W exist. Refer to \cref{sec:related-work} for more details.
}
\vskip 0.15in
{\footnotesize
\begin{tabular}{l c c c c c c} 
    \toprule
    
    & \rot{No 3D supervision} 
    & \rot{Pre-trained prior} 
    & \rot{Geometric consistency} 
    & \rot{In-the-wild scene modeling} 
    & \rotp{Underlying representation} 
    & \rotp{Task} \\ \midrule

    NFD \citep{3D-nf-gen-triplane-diffusion}
    & \xmark 
    & \xmark 
    & \cmark 
    & \xmark 
    & Tri-Planes 
    & \multirow{4}{*}{OG} \\
    
    3DGen \citep{3dgen}
    & \xmark 
    & \xmark 
    & \cmark 
    & \xmark 
    & Tri-Planes \\
    
    Latent-NeRF \citep{latentnerf}
    & \cmark 
    & \cmark 
    & \cmark 
    & \xmark 
    & NeRF \\
    
    DreamFusion \citep{dreamfusion}
    & \cmark 
    & \cmark 
    & \cmark 
    & \xmark 
    & NeRF \\ \midrule

    NeRF \citep{nerf}
    & \cmark 
    & \xmark 
    & \cmark 
    & \xmark 
    & NeRF 
    & \multirow{7}{*}{NVS} \\
    
    RealFusion \citep{realfusion}
    & \cmark 
    & \cmark 
    & \cmark 
    & \xmark 
    & NeRF \\
    
    DiffusioNeRF \citep{diffusionerf}
    & \cmark 
    & \xmark 
    & \cmark 
    & \xmark 
    & NeRF \\
    
    NerfDiff \citep{nerfdiff}
    & \xmark 
    & \cmark 
    & \cmark 
    & \xmark 
    & Tri-Planes \\
    
    3DiM \citep{nvs-diffusion-models}
    & \xmark 
    & \xmark 
    & \cmark 
    & \xmark 
    & --- \\
    
    Zero-1-to-3 \citep{zero-1-to-3}
    & \xmark 
    & \cmark 
    & \xmark 
    & \xmark 
    & --- \\ \midrule
    
    NeRF-W \citep{nerf-w}
    & \cmark 
    & \xmark 
    & \cmark 
    & \cmark 
    & NeRF 
    & \multirow{5}{*}{NVS-W$^{\dagger}$} \\

    WildGaussians \citep{wildgaussians}
    & \cmark
    & \xmark
    & \cmark
    & \cmark
    & 3DGS \\

    Splatfacto-W \citep{splatfacto-w}
    & \cmark
    & \xmark
    & \cmark
    & \cmark
    & 3DGS \\

    K-Planes \citep{kplanes}
    & \cmark 
    & \xmark 
    & \cmark 
    & \cmark 
    & K-Planes \\
    
    RefinedFields (ours)
    & \cmark 
    & \cmark 
    & \cmark 
    & \cmark 
    & K-Planes \\ \bottomrule
    
\end{tabular}
}
\label{table:comparison}
\end{table}

\paragraph{Neural representations in-the-wild.}
Subsequent to NeRFs, several techniques emerged to extend the NeRF setup to ``in-the-wild'' unconstrained photo collections \citep[Phototourism]{phototourism} plagued by illumination variations and transient occluders.
This added variability makes learning a scene particularly challenging, as surfaces can exhibit significant visual disparities across views.
NeRF-W \citep{nerf-w} addresses the challenge of novel view synthesis in-the-wild (NVS-W) by modeling scene lighting through appearance embeddings, and transient occluders through an additional transient head. 
\mbox{Splatfacto-W} \citep{splatfacto-w} proposes a Nerfstudio \citep{nerfstudio} implementation of 3D Gaussian Splatting \citep[3DGS]{gaussian-splatting} that extends standard 3DGS to support in-the-wild scene modeling via appearance embeddings.
WildGaussians \citep{wildgaussians} also proposes using 3DGS with appearance embeddings for in-the-wild scene modeling, but additionally predicts uncertainty masks to exclude transient occluders from the loss computation.
It is important to note that although this work adopts a pre-trained network, it is solely used for the computation of the loss masks on the data, and not as a prior on the scene representation.
This approach is orthogonal to our approach of leveraging a prior to directly improve upon a specific scene representations, as it acts on the data and not the scene representation itself.
\citet{kplanes} present K-Planes, which modify and extend Tri-Planes \citep{eg3d} to in-the-wild scenes thanks to learnable appearance embeddings, similarly to \citet{nerf-w}. 
Our work aims to improve upon K-Planes representations by extending their training beyond closed-world setups using pre-trained networks.
Note that other works also tackle NVS-W:
\citet{ha-nerf} model scene lighting through appearance embeddings and transient occluders through transient embeddings,
\citet{cr-nerf} leverage interactive information across rays to mimic the perception of humans in-the-wild,
and \citet{nerf-hugs} separate static and transient components by utilizing heuristics-guided segmentation.
However, these works limit their training and evaluations to downscaled versions of Phototourism, which makes their results not directly comparable with ours.

\paragraph{Priors in neural representations.}
The integration of pre-trained priors for downstream tasks has emerged as a prominent trend, as they enable the effective incorporation of extrinsic knowledge into diverse applications.
For neural representations, priors have been utilized for few-shot scene modeling \citep{pixelnerf, dietnerf, nerdi, cat3d} as well as generative tasks \citep{3D-nf-gen-triplane-diffusion,dreamfusion,nvs-diffusion-models, cat3d} for object generation and novel view synthesis.
In this realm, denoising diffusion probabilistic models \citep{ddpm, stable-diffusion} have recently gained particular attention for their application as plug-and-play priors \citep{diffusion-prior}.
They have been utilized in various domains such as super-resolution \citep{sd-for-super-resolution} and more specifically novel view synthesis.
\citet{zero-1-to-3} fine-tune Stable Diffusion \citep{stable-diffusion}, a pre-trained latent diffusion model for 2D images, to learn camera controls over a 3D dataset and thus performing NVS by generalizing to other objects.
These results hold paramount value, as they highlight the rich 3D prior learned by Stable Diffusion, even though it has only been trained on 2D images.
This however comes with geometric inconsistency issues across views, as a pre-trained model alone has no explicit multi-view geometric constraints.
In this work, we aim to leverage a pre-trained model to enhance a volumetric scene representation, which inherently adheres to consistency constraints, hence mitigating multi-view consistency issues.
The utilization of such priors to enhance in-the-wild scene representation has been an unexplored area of research.

\section{Method}
\begin{figure*}[t]
    \centering
    \includegraphics[width=\textwidth]{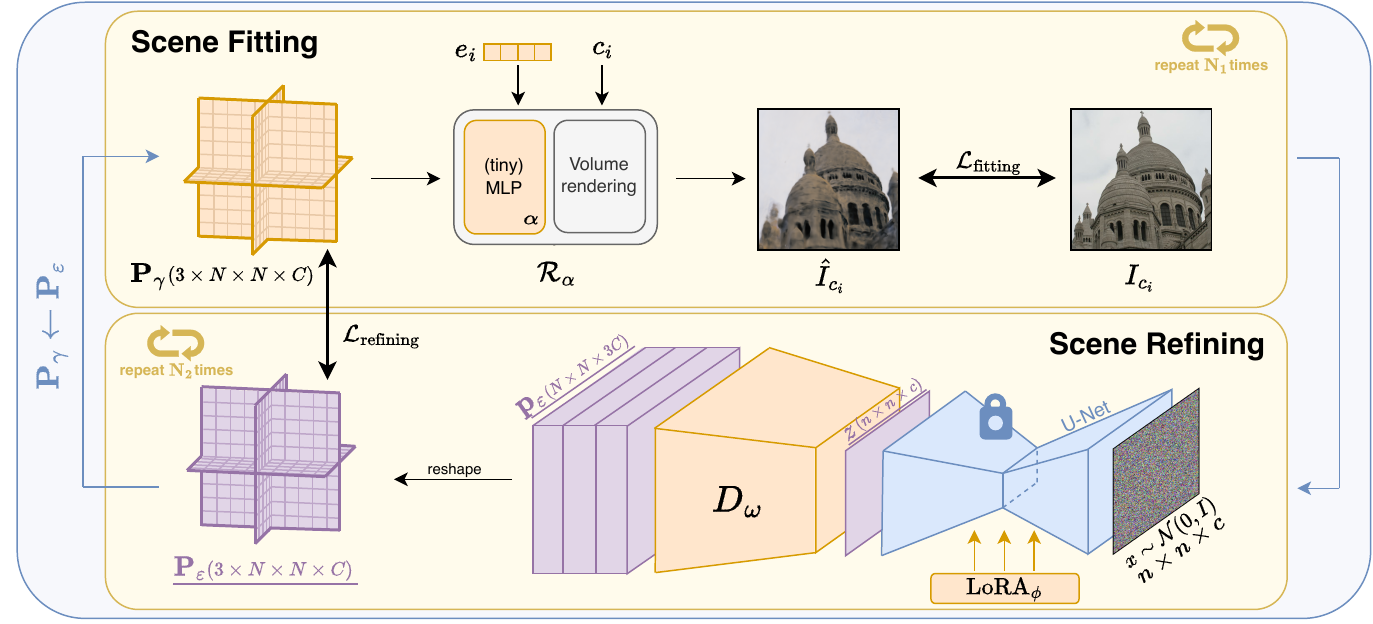}
    \caption{\textbf{Scene learning procedure.} The K-Planes $\*P_\gamma$, the MLP with trainable parameters $\alpha$, and the appearance embeddings $e_i$ are learned during scene fitting. The LoRA parameters $\phi$ as well as the decoder $D_w$ are learned during scene refining. The pre-trained U-Net is frozen. Assets in \textcolor{violet}{\underline{violet and underlined}} are intermediate results. At each iteration, new planes $\*P_\varepsilon$ are inferred and assigned to $\*P_\gamma$, which are then corrected by scene fitting.}
    \label{fig:method-training}
\end{figure*}

To guide the optimization of planar scene representations with extrinsic signals, we learn a scene through two alternating stages, as illustrated in \cref{fig:method-training}.
\emph{Scene fitting} optimizes our K-Planes representation $\*P_\gamma$ to reproduce the images in the training set, as traditionally done in neural rendering techniques.
\mbox{\emph{Scene refining}} finetunes a pre-trained network to this K-Planes representation, and then infers a new one $\*P_\varepsilon$, which will subsequently be corrected by scene fitting. 
The main idea behind this is that we use our 3D implicit model $\*P_\gamma$ for optimizing the scene to the available information in the training set and adhere to essential geometric constraints, and then project this scene representation on the set of scenes inferable by the pre-trained network, making it closer to natural images.
In this section, we detail each stage and elucidate the intuition behind our method.

\subsection{Scene Fitting}
\label{sec:scene-fitting}
The goal at this stage is to fit a scene, adhering to pre-defined geometric constraints, from posed RGB images. 
To fit the scene, we adopt the K-Planes representation \citep{kplanes}.
As such, this stage corresponds to optimizing K-Planes to fit a scene, from which we adapt the code.

K-Planes are compact 3D model representations applicable to static scenes, ``in-the-wild'' scenes (scenes with varying appearances), and dynamic scenes.
These models allow for fast training and rendering, while maintaining low-memory usage. K-Planes model a d-dimensional scene with $k = \binom{d}{2}$ planes, which represent the combinations of every pair of dimensions.
This structure makes K-Planes compatible with a multitude of neural network architectures, and more particularly image-specialized network architectures. 
This enables K-Planes inference by minimally tweaking image architectures.
For a static 3D scene, $k = 3$ and the planes represent the $xy, xz, $ and $yz$ planes.
These planes, each of size $N \times N \times C$, encapsulate features representing the density and view-dependent colors of the scene. 

The K-Planes model $\*P_\gamma$ is originally randomly initialized. The first goal of scene fitting is then to correct this random initialization to fit the training set.
Note that the first iteration of scene fitting is especially particular, since it is starting with a randomly initialized scene, as opposed to a \emph{proposed} scene, as we describe in \cref{sec:scene-refining}.

To render the 3D scene from K-Planes, as done by \citet{nerf} and \citet{kplanes}, we cast rays from the desired camera position through the coordinate space of the scene, on which we sample 3D points.
We decode the corresponding RGB color for each 3D point $\mathbf{q} = (i, j, k)$ by normalizing it to $[0,N)$ and projecting it onto the $k=3$ planes, denoted as $\mathbf{P}_\gamma^{(xy)}, \mathbf{P}_\gamma^{(xz)}, \mathbf{P}_\gamma^{(yz)}$:
\begin{equation}
    f^{(h)}(\mathbf{q}) = \psi (\mathbf{P}_\gamma^{(h)}, \pi^{(h)}(\mathbf{q}))~,
\end{equation}
where $h \in \mathbf{H} = \{xy, xz, yz\}$, $\pi^{(h)}(\mathbf{q})$ projects $\mathbf{q}$ onto $\mathbf{P}_\gamma^{(h)}$, and $\psi$ denotes bilinear interpolation on a regular 2D grid.

These features are then aggregated using the Hadamard product to produce a single feature vector of size $M$:
\begin{equation}
    f(\mathbf{q}) = \prod_{h \in \mathbf{H}} f^{(h)}(\mathbf{q})~.
\end{equation}
To decode these features, we adopt the hybrid formulation of K-Planes \citep{kplanes}.
Two small Multi-Layer Perceptrons (MLPs), $g_\sigma$ and $g_{RGB}$, map the aggregated features as follows:
\begin{equation}
\begin{split}
    \sigma(\mathbf{q}), \hat{f}(\mathbf{q}) &= g_\sigma(f(\mathbf{q}))~,\\
    c(\mathbf{q}, \mathbf{d}) &= g_{\mathrm{RGB}}(\hat{f}(\mathbf{q}), \gamma(\mathbf{d}))~,
\end{split}
\end{equation}
where $\gamma(p) = (\sin(2^0 \pi p), \cos(2^0 \pi p), \ldots, \sin(2^{L-1}\pi p),$ $\cos(2^{L-1} \pi p))$ is the positional embedding of $p$. $g_\sigma$ maps the K-Planes features into density $\sigma$ and additional features $\hat{f}$. Subsequently, $g_{\mathrm{RGB}}$ maps $\hat{f}$ and the positionally-encoded view directions $\gamma(\mathbf{d})$ into view-dependent  RGB colors. This enforces densities to be independent of view directions.

These decoded RGB colors are then used to render the final image thanks to ray marching and integrals from classical volume rendering \citep{volume-rendering}, that are practically estimated using quadrature:
\begin{equation}
\begin{split}
    \hat{C}(\*r) &= \sum_{i=1}^{N} T_i (1 - \exp(-\sigma_i \delta_i)) c_i~, \\
    T_i &= \exp \left ( - \sum_{j=1}^{i-1} \sigma_j \delta_j \right )~,
\end{split}
\end{equation}
where $\hat{C}(\*r)$ is the expected color, $T_i$ is the accumulated transmittance along the ray, and  $\delta_i = t_{i+1} - t_{i}$ is the distance between adjacent samples.

\subsection{Scene Refining}
\label{sec:scene-refining}
\begin{figure*}[t]
    \centering
    \includegraphics[width=\textwidth]{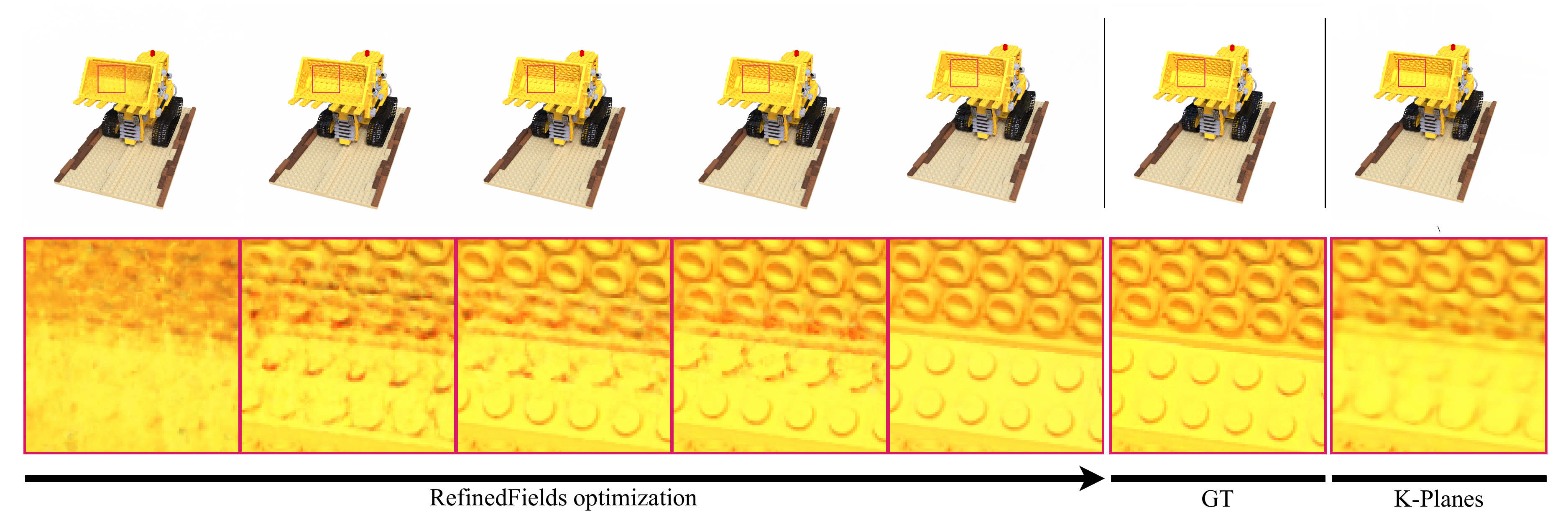}
    \caption{\textbf{Case study.} Qualitative results on the Lego scene from the NeRF synthetic dataset \citep{nerf} showcasing the optimization progression on RefinedFields, and a comparison with the ground truth and K-Planes. The training set is constrained to 50\% of its initial size for both RefinedFields and K-Planes. RefinedFields refines the K-Planes representation enabling the proper reconstruction of details in the scene. At the end of optimization, the Mean Squared Error (MSE) for RefinedFields is $3.46 \times 10^{-4}$, while the one for K-Planes is $4.36 \times 10^{-4}$.}
    \label{fig:lego-progress}
\end{figure*}

This section presents the core of our method, which consists of proposing better optimization initializations for the scene fitting stage.
Given a fitted scene representation $\*P_\gamma$, this stage consists of learning this fitted implicit representation and proposing a new \emph{refined} representation $\*P_\varepsilon$.
Formally, this stage consists of projecting our K-Planes $\*P_\gamma$ on the set $\mathbb{Q}$ of K-Planes inferable by a low-rank fine-tuning of the pre-trained model:
\begin{equation}
    \*P_\varepsilon = \argmin_{P \in \mathbb{Q}}~\lVert \*P_\gamma - P \rVert_2^2~.
\end{equation}
As K-Planes feature channels show similar structure to images (\cref{x:inspection}), this projection pushes the K-Planes to be even more similar in structure to real images, more particularly to orthogonal projections of the scene on the planes. 
\cref{fig:inspection-0,fig:inspection-1,fig:inspection-2} illustrate a comparison between feature planes at the end of our optimization and those of standard K-Planes.
RefinedFields leads to feature planes exhibiting sharper details, which ultimately lead to more refined details in scene renderings, as proven by our experiments (\cref{fig:lego-progress,tab:results-itw}).

To provide scene refining with a rich prior, we employ a large-scale pre-trained latent diffusion model, as these networks exhibit great performances as priors for downstream tasks, and share similar properties to our planar representation, both in terms of shape and distribution (\cref{x:inspection}).
More particularly, we adopt Stable Diffusion \citep[SD]{stable-diffusion} for its proven performances for downstream 3D \citep{zero-1-to-3} and 2D \citep{sd-for-super-resolution} tasks.
Thus, we integrate the U-Net $\mathbf{SD}_\phi$ and the decoder $\mathbf{D}_\omega$ into our pipeline, and treat the K-Planes as $3C$-channel $N \times N$ images. We also replace the last layer of the decoder $\*D_\omega$ with a randomly initialized convolutional layer (with no bias), to take into account the shape of the K-Planes. 
We then fine-tune the pre-trained model using the fitted K-Planes $\*P_\gamma$, and infer refined K-Planes $\*P_\varepsilon = \*D_w(\mathbf{SD}_\phi (x))$ where $x \sim \mathcal{N}(0, I)$ is sampled once at the beginning of the training.
Note that this is different from the multi-step generation process of diffusion model inference, as we only apply the inference at the last time-step of our diffusion model.
This is key as our goal here is not to learn distributions over scenes and sample them for generation, but to adapt the pre-trained network and leverage the information already learned within its weights to infer K-Planes closest to representing the scene at hand. 

To achieve the fine-tuning of our pre-trained network, a significant challenge presents itself: due to the sheer size of Stable Diffusion, it would be too costly to fine-tune all of its trainable parameters.
Moreover, as we only want to modulate priors embedded into the pre-trained network, we look for an alternative to doing full fine-tuning.
To circumvent these constraints, we adopt Low-Rank Adaptation \citep[LoRA]{lora}, a simple yet effective parameter-efficient fine-tuning method that has proven great transfer capabilities across modalities and tasks \citep{rl-ft-lora, lee2023platypus, expressive-lora}.
LoRA's relatively minimal design works directly over weight tensors, which means that it can be seamlessly applied to most model architectures.
Furthermore, LoRA does not add any additional cost at inference, thanks to its structural re-parameterization design.
To achieve this, \citet{lora} inject trainable low-rank decomposition matrices into each layer of a frozen pre-trained model.
Let $\mathbf{W}_0$, $\mathbf{b}_0$ be the frozen pre-trained weights and biases, and $x$ be the input.
Fine-tuning a frozen linear layer $f(x) = \mathbf{W}_0 x + \mathbf{b}_0$ comes down to learning the low-rank decomposition weights $\Delta \mathbf{W} = \*B \*A$:
\begin{equation}
    f(x) = (\mathbf{W}_0 + \Delta \mathbf{W})x + \mathbf{b}_0
\end{equation}
where $\mathbf{W}_0, \Delta \mathbf{W} \in \mathbb{R}^{d \times k}$; $\*B \in \mathbb{R}^{d \times r}$; $\*A \in \mathbb{R}^{r \times k}$; and the rank $r \ll \min(d,k)$.

Thus, to implement scene refining, we fine-tune the LoRA parameters $\phi$ modulating the pre-trained U-Net, as well as the decoder's parameters $\omega$,  on the fitted scene $\*P_\gamma$.
Subsequently, we query the U-Net with Gaussian noise $x$, decode its intermediary output latent $z$ with $\*D_\omega$, and infer $\*p_\varepsilon$ that is reshaped into a new \emph{refined} scene $\*P_\varepsilon$.
Finally, $\*P_\varepsilon$ is proposed to \emph{scene fitting} as an improved initialization to be optimized.
For an in-depth inspection of the feature planes, we refer the reader to \cref{x:inspection}. 

\begin{algorithm}[tb]
   \caption{Alternating training algorithm.}
   \label{alg:alternate-training}
\begin{algorithmic}[1]
   \STATE {\bfseries Input:} $N_{\mathrm{epochs}}$, $N_1$, $N_2$, $N$, $C$, $n$, $c$, $\mathcal{I} = \{I_{c_i}, c_i\}$, $\mathcal{R}_\alpha$, $\*D_w$,  $\mathbf{SD}_\phi$, $\text{optimizer}$
   \STATE $x \gets \text{standard-gaussian}(n, n, c)$
   \STATE $\*P_\gamma \gets \text{standard-gaussian}(N, N, 3C)$
   \FOR{$N_{\mathrm{epochs}}$ steps}
       \STATE \textcolor{gray}{\footnotesize \textit{// scene fitting}}
       \FOR{$N_{1}$ steps}
       \STATE $\gamma, \alpha \leftarrow \text{optimizer.step}(\mathcal{L}_{\mathrm{fitting}}(\*P_\gamma, \mathcal{I}))$\;
       \ENDFOR
       \STATE \textcolor{gray}{\footnotesize \textit{// scene refining}}
       \FOR{$N_{2}$ steps}
       \STATE $\*P_\varepsilon \gets \*D_w (\mathbf{SD}_\phi (x))$
       \STATE $\omega, \phi \leftarrow \text{optimizer.step}(\mathcal{L}_\mathrm{refining}(\*P_\varepsilon, \*P_\gamma))$
       \ENDFOR
       \STATE $\*P_\gamma \gets \*P_\varepsilon$\;
   \ENDFOR
\end{algorithmic}
\end{algorithm}
\subsection{Training}
We define an alternating training procedure rotating between scene fitting and scene refining, as described above, and as illustrated in \cref{fig:method-training}.

For \emph{scene fitting}, we train the K-Planes model as proposed by \citet{kplanes}.
We use spatial total variation regularization to encourage smooth gradients. This is applied over all the spatial dimensions of each plane in the representation: 
\begin{equation}
    \mathcal{L}_{\mathrm{TV}} (\*P) = \frac{1}{\vert C \vert N^2} \sum_{c,i,j} (\lVert \*P_c^{i,j} - \*P_c^{i-1,j} \rVert_2^2
    + \lVert \*P_c^{i,j} - \*P_c^{i,j-1} \rVert_2^2)~.
\end{equation}
For scenes with varying lighting conditions (e.g.\ \emph{in-the-wild} scenes as in the Phototourism dataset \citep{phototourism}), an $M$-dimensional appearance vector $e_i$ is additionally optimized for each image.
This vector is then passed as input to the MLP color decoder $g_{RGB}$ at the rendering step $\mathcal{R}_\alpha$. 
Hence, the training objective for scene fitting is written as: 
\begin{equation} 
    \displaystyle\min_{\alpha, \gamma} \mathcal{L}_{\mathrm{fitting}} \triangleq \lVert \mathcal{R}_\alpha(\*P_\gamma, \*C) - I_{\*C} \rVert_2^2 + \lambda_{\mathrm{TV}}\mathcal{L}_{\mathrm{TV}} (\*P_\gamma)~,
\end{equation}
where $\mathcal{R}_\alpha$ represents the K-Planes rendering procedure (i.e.\ ray marching, feature decoding via a small MLP with trainable parameters $\alpha$, and volume rendering), $\*P_\gamma$ are the K-Planes with trainable parameters $\gamma$ and $I_{\*C}$ is a ground truth RGB image with camera position $\*C$.

As for the \emph{scene refining} phase, we optimize the decoder parameters $w$ as well as the LoRA parameters $\phi$, modulating the frozen U-Net weights, on the fitted scene $\*P_\gamma$.
Thus, the fine-tuning objective for \emph{scene refining} is written as:
\begin{equation}
    \displaystyle\min_{w, \phi} \mathcal{L}_{\mathrm{refining}} \triangleq \lVert \*D_w (\mathbf{SD}_\phi (x)) - \*P_\gamma \rVert_2^2~,
\end{equation}
where $x \sim \mathcal{N}(0,I)$ is fixed during scene refining, $\mathbf{SD}_\phi$ is the frozen Stable Diffusion model modulated by LoRA with trainable parameters $\phi$, and $\*D_w$ is the latent K-Planes decoder. After this optimization, $\*P_\gamma$ is reassigned as $\*D_w (\mathbf{SD}_\phi (x))$ and passed to scene fitting.
Note that, thanks to the alternating nature of our training and the absence of bias in the decoder's convolutional layers, this optimization does not overfit the model to produce exactly $\*P_\gamma$, which is key as it would lead to resuming scene fitting from exactly the same point.

At the end of the alternating training procedure, we save the refined and corrected representation $\*P_\gamma$ for rendering and testing. We refer the reader to \Cref{alg:alternate-training} for an overview of our training procedure.

\section{Experiments}
We start by assessing RefinedFields via an experiment on a case study. 
We then evaluate RefinedFields on synthetic scenes \citep{nerf} and real-world Phototourism \citep{phototourism} scenes, where we showcase the improvements our method exhibits relative to our K-Planes base representation.
Quantitative results can be found in \cref{tab:results-synth,tab:results-itw}, where we report for each experiment the Peak Signal-to-Noise Ratio (PSNR) for pixel-level similarity, the Structural Similarity Index Measure (SSIM) for structural-level similarity, and the Learned Perceptual Image Patch Similarity \citep[LPIPS]{lpips} for perceptual similarity. 
RefinedFields demonstrates an improved performance compared to K-Planes on the task of novel view synthesis.
For a further look, experimental details including hyperparameters and more dataset details can be found in Appendix \ref{x:exp-details}. 
Additional qualitative results on synthetic and real-world scenes are available in Appendix \ref{x:results}. 

\subsection{Datasets}
We evaluate our method similarly to prior work \citep{nerf, nerf-w, kplanes} in novel view synthesis, by adopting the \emph{Real Synthetic $\mathit{360^{\circ}}$} dataset \citep{nerf} for synthetic scenes and the same three scenes of cultural monuments from the Phototourism dataset \citep{phototourism} for real-world scenes: \emph{Brandenburg Gate}, \emph{Sacré Coeur}, and \emph{Trevi Fountain}. 
Additional dataset details can be found in \cref{x:dataset-details}.

\begin{table*}[t]
    \small
    \centering
    \caption{\textbf{Quantitative results. } Results on static synthetic scenes \citep{nerf}. The \textbf{bold} and \underline{underlined} entries respectively indicate the best and second-best results. Dashes denote values that were not reported in prior work. Our method outperforms K-Planes, our main baseline, on the task of novel view synthesis for synthetic scenes.}
    \vskip 0.15in
    \resizebox{\textwidth}{!}{
        \begin{tabular}{l c c c c c c c c c c}
            \toprule
             & \multicolumn{9}{c}{PSNR ($\uparrow$)}\\
             \cmidrule{2-9} 
             & Chair & Drums & Ficus & Hotdog & Lego & Materials & Mic & Ship & & Mean\\
            \midrule
            NeRF \citep{nerf} & 33.00 & 25.01 & 30.13 & 36.18 & 32.54 & 29.62 & 32.91 & 28.65 & & 31.00\\
            TensoRF \citep{tensorf} & \underline{35.76} & \underline{26.01} & \textbf{33.99} & \textbf{37.41} & \underline{36.46} & \textbf{30.12} & 34.61 & 30.77 & & \underline{33.14}\\
            Plenoxels \citep{plenoxels} & 33.98 & 25.35 & 31.83 & 36.43 & 34.10 & 29.14 & 33.26 & 29.62 & & 31.71\\
            INGP \citep{instantngp} & 35.00 & \textbf{26.02} & \underline{33.51} & \underline{37.40} & 36.39 & \underline{29.78} & \textbf{36.22} & \underline{31.10} & & \textbf{33.18}\\
            \midrule
            K-Planes \citep{kplanes} & 34.98 & 25.68 & 31.44 & 36.75 & 35.81 & 29.48 & 34.10 & 30.76 & & 32.37\\
            K-Planes-SS \citep{kplanes} & 33.61 & 25.27 & 30.92 & 35.88 & 35.09 & 28.83 & 33.01 & 30.04 & & 31.58\\
            RefinedFields (ours) & \textbf{35.77} & 25.94 & 32.45 & 37.08 & \textbf{36.47} & 29.39 & \underline{34.77} & \textbf{31.41} & & 32.91\\
            \bottomrule
        \end{tabular}
    }
    \resizebox{\textwidth}{!}{
        \begin{tabular}{l c c c c c c c c c c}
            \toprule
             & \multicolumn{9}{c}{SSIM ($\uparrow$)}\\
             \cmidrule{2-9} 
             & Chair & Drums & Ficus & Hotdog & Lego & Materials & Mic & Ship & & Mean\\
            \midrule
            NeRF \citep{nerf} & 0.967 & 0.925 & 0.964 & 0.974 & 0.961 & 0.949 & 0.980 & 0.856 & & 0.947\\
            TensoRF \citep{tensorf} & \textbf{0.985} & \underline{0.937} & \textbf{0.982} & \textbf{0.982} & 0.983 & \textbf{0.952} & \underline{0.988} & 0.895 & & \textbf{0.963}\\
            Plenoxels \citep{plenoxels} & 0.977 & 0.933 & 0.976 & 0.980 & 0.975 & 0.949 & 0.985 & 0.890 & & 0.958\\
            INGP \citep{instantngp} & --- & --- & --- & --- & --- & --- & --- & --- & & ---\\
            \midrule
            K-Planes \citep{kplanes} & \underline{0.983} & \textbf{0.938} & 0.975 & \textbf{0.982} & \underline{0.982} & \underline{0.950} & \underline{0.988} & \underline{0.897} & & \underline{0.962}\\
            K-Planes-SS \citep{kplanes} & 0.974 & 0.932 & 0.971 & 0.977 & 0.978 & 0.943 & 0.983 & 0.887 & & 0.956\\
            RefinedFields (ours) & \textbf{0.985} & \underline{0.937} & \underline{0.980} & \underline{0.981} & \textbf{0.984} & 0.945 & \textbf{0.989} & \textbf{0.903} & & \textbf{0.963}\\
            \bottomrule
        \end{tabular}
    }
\label{tab:results-synth}
\end{table*}

\subsection{Implementation Details}
\label{sec:imp-details}
For a fair comparison, we take similar experimental settings in scene fitting to \citet{kplanes}.
However, due to the nature of our scene refining pipeline, we limit the implementation in our case to a single-scale K-Planes of $512 \times 512$ resolution, in contrast to the multi-scale approach taken by \citet{kplanes} where $N \in \{64, 128,256,512\}$.
The number of channels in each plane remains the same ($C=32$).
Moreover, throughout all the experiments, we consider the hybrid implementation of K-Planes, where plane features are decoded into colors and densities by a small MLP.
As for the scene refining pipeline, we apply no modification to the U-Net in Stable Diffusion.
Yet, we replace the last layer of the decoder $D_\omega$ with a new convolutional layer (without bias) to account for the shape of the K-Planes.
Hence, the dimensions used in scene refining (\cref{fig:method-training}) are: $N=512$, $C=32$, $n=64$, and $c=4$.
For an in-depth look at our frameworks and hyperparameter settings, we refer the reader to \cref{x:frameworks-details,x:hyperparameter-details}.

\subsection{Evaluations}

\paragraph{Baselines.}
For both synthetic and in-the-wild scenes, we primarily compare our method to our K-Planes baseline, highlighting the improvements it brings to planar scene representations.
Note that, for a fair comparison, and to assess the added value of our refining pipeline with respect to our base representation (\cref{sec:imp-details}), we also include a single-scale ablation of K-Planes with $N = 512$ (dubbed K-Planes-SS).
In order to provide a more comprehensive perspective, we also illustrate the NVS performances of other recent works that do not employ planar scene representations for synthetic \citep{nerf,tensorf,plenoxels,instantngp} and in-the-wild \citep{nerf, nerf-w, splatfacto-w, wildgaussians} scenes.

\paragraph{Comparisons.}
We start by testing our method against K-Planes on a \textbf{case study} consisting of the Lego scene from the NeRF synthetic dataset. 
Here, we train both methods on half of the training set as to deliberately produce a lower-quality fitted scene. 
As illustrated in \cref{fig:lego-progress} our method refines K-Planes and exhibits better quantitative and qualitative results thanks to our scene refining pipeline. 

\begin{figure*}[t]
    \centering
    \includegraphics[width=\textwidth]{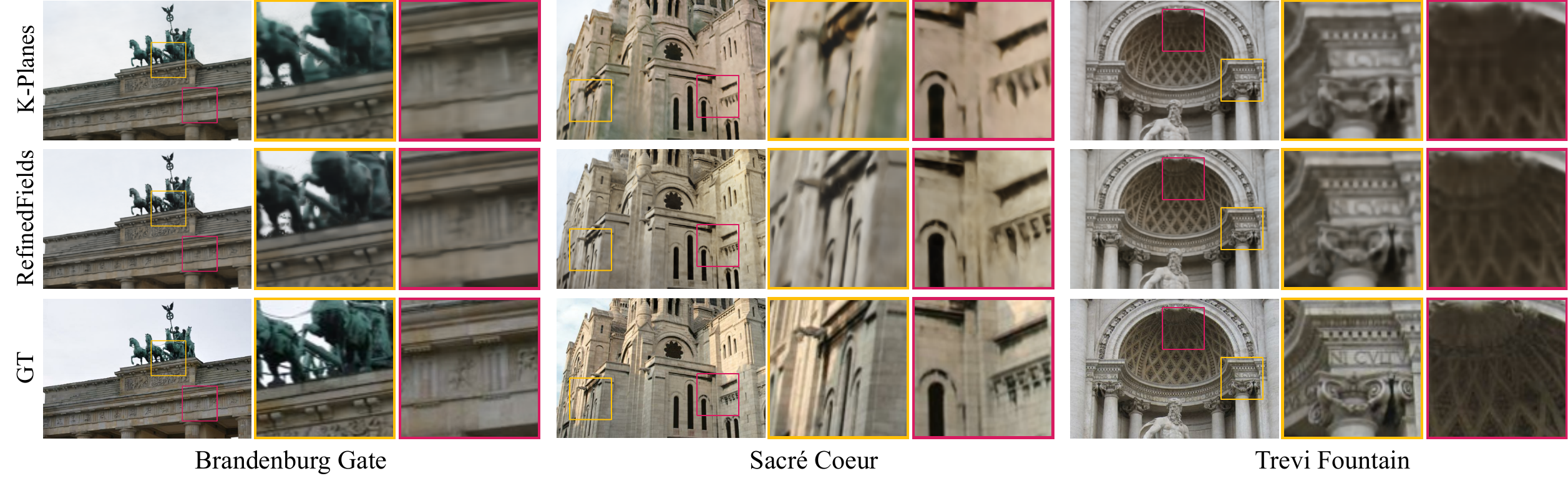}
    \caption{\textbf{Qualitative results.} Results on three scenes from Phototourism \citep{phototourism}. Our method refines K-Planes and leads to richer and finer details in scene renderings.}
    \label{fig:microscope}
\end{figure*}

We then apply our method to learn synthetic and in-the-wild scenes. 
In this case, RefinedFields improves upon not only K-Planes-SS but also K-Planes on the task of NVS (\cref{tab:results-synth,tab:results-itw}). 
As a result, it enhances the performance of planar scene representations and brings them closer to recent works, which highlights the value of scene refining.
Particularly, for synthetic scenes, RefinedFields improves upon K-Planes and sometimes even outperforms other state-of-the-art methods.
For in-the-wild scenes, RefinedFields also improves K-Planes performances in a notable way. 
However, recent state-of-the-art methods based on Gaussian Splatting architectures \citep{gaussian-splatting} continue to outperform the refined planar representations in terms of overall quality.
\cref{fig:storefront,fig:microscope} show qualitative comparisons of RefinedFields with K-Planes, showing the visual improvements brought by our refining pipeline, which brings finer details to monuments in the Phototourism scenes.
Further qualitative results on synthetic and in-the-wild scenes can be found in \cref{x:results}.
We also present an inspection of K-Planes features learned by our method and K-Planes in \cref{x:inspection}.

It is important to note that we do not compare our NVS metrics with other recent works \citep{ha-nerf, cr-nerf, gs-w}.
This is because these methods are trained and evaluated on downscaled versions of the Phototourism dataset, which reduces the prominence of fine details in the ground truth for monument structures.
As their NVS metrics are computed relative to these ground truth, their results are not comparable to ours.
Although other works do compare NVS metrics of various methods across different resolutions, we refrain from this practice as it is not an accurate comparison.

\begin{table*}[t]
    \small
    \centering
    \caption{
    \textbf{Quantitative results. } Results on three real-world datasets from Phototourism \citep{phototourism}. 
    Our method shows notable improvements compared to K-Planes on the task of NVS-W. \\
    $^{\dagger}$Results from public implementation \citep{nerfw-public} reproduced by \citet{kplanes}.
    }
    \begin{adjustbox}{width=\textwidth}
    \begin{tabular}{l c c c c c c c c c c c}
        \toprule
         & \multicolumn{3}{c}{Brandenburg Gate} & 
         & \multicolumn{3}{c}{Sacré Coeur} & 
         & \multicolumn{3}{c}{Trevi Fountain}\\
         \cmidrule{2-4}
         \cmidrule{6-8}
         \cmidrule{10-12}
         & PSNR ($\uparrow$) & SSIM ($\uparrow$) & LPIPS ($\downarrow$) & 
         & PSNR ($\uparrow$) & SSIM ($\uparrow$) & LPIPS ($\downarrow$) & 
         & PSNR ($\uparrow$) & SSIM ($\uparrow$) & LPIPS ($\downarrow$)\\
        \midrule
        
        NeRF & 18.90 & 0.8159 & 0.231 & & 15.60 & 0.7155 & 0.291 & & 16.14 & 0.6007 & 0.366 \\
        NeRF-W$^{\dagger}$ & 21.32 & --- & --- & & 19.17 & --- & --- & & 18.61 & --- & --- \\
        Splatfacto-W & 26.87 & \textbf{0.9320} & \textbf{0.124} & & 22.53 & \textbf{0.8760} & \textbf{0.158} & & 22.66 & \textbf{0.7690} & \textbf{0.224} \\
        WildGaussians & \textbf{27.77} & 0.9270 & 0.133 & & \textbf{22.56} & 0.8590 & 0.177 & & \textbf{23.63} & 0.7660 & 0.228 \\ \midrule
        K-Planes & 25.49 & 0.8785 & 0.224 & & 20.61 & 0.7735 & 0.265 & & 22.67 & 0.7139 & 0.317 \\
        K-Planes-SS & 24.48 & 0.8629 & 0.242 & & 19.86 & 0.7419 & 0.312 & & 21.30 & 0.6627 & 0.355 \\
        RefinedFields-noFinetuning & 25.39 & 0.8834 & 0.206 & & 21.41 & 0.8059 & 0.239 & & 22.54 & 0.7324 & 0.291 \\
        RefinedFields-noPrior & 25.42 & 0.8822 & 0.214 & & 21.17 & 0.7978 & 0.248 & & 22.16 & 0.7251 & 0.291 \\
        \textbf{RefinedFields (ours)} & 26.64 & 0.8869 & 0.206 & & 22.26 & 0.8176 & 0.228 & & 23.42 & 0.7379 & 0.284 \\
        \bottomrule
    \end{tabular}
    \end{adjustbox}
\label{tab:results-itw}
\end{table*}

As presented, RefinedFields utilizes an alternating training procedure and a pre-trained prior to refine scene representations, leading to richer details in rendered images. 
While our method demonstrates promising results, this however comes with a training time increase as compared to our base representation, as our K-Planes feature projection via alternating training leads to the repeated fine-tuning of both scene fitting and scene refining pipelines, which takes overall around 80 hours on a single NVIDIA A100 GPU. 
We leave the optimization of training time for future work.

\subsection{Ablations}
To justify our choices and explore further, we compare our in-the-wild results (\cref{tab:results-itw}) to results from two main ablations of our method. 
\textbf{RefinedFields-noFinetuning} is a variation of our method without LoRA fine-tuning. Here, we consider the same exact pipeline (frozen U-Net, same decoder configuration), except that we don't modulate the weights of the frozen U-Net with LoRA. This means that the prior is kept intact and no fine-tuning is done. 
This is to assess the role that LoRA finetuning of the pre-trained prior plays in our pipeline.
Note that the decoder is still fine-tuned in this ablation, as additionally freezing it would lead to a random constant optimization initialization $\mathbf{P}_\varepsilon$, which would defeat the purpose of the refining stage.
\textbf{RefinedFields-noPrior} ablates the prior of Stable Diffusion by randomly re-initializing all U-Net weights while leaving all other elements of the scene-refining pipeline intact. This ablation is done to evaluate the importance of the prior, and to verify that the observed refinements are not entirely resulting from alternate training.
Note that ablating the entire scene refining pipeline leads back to the \textbf{K-Planes-SS} setting.
As illustrated in \cref{tab:results-itw}, we consistently obtain worse results during the ablation study as compared to our full model, thus demonstrating the value of the pre-trained prior and of LoRA finetuning. 
This proves the importance of extending scene learning beyond closed-world settings, as our K-Planes projection is done via an optimization conditioning coming from a large image prior that a reasonably sized training set cannot fully capture, especially in-the-wild.

\section{Conclusion}
In this paper, we introduce RefinedFields, a method that refines K-Planes representations by using a pre-trained prior and an alternate training procedure.
Extensive experiments show that RefinedFields exhibits notable improvements on the task of novel view synthesis compared to its K-Planes baseline.
In concluding this study, several avenues of future work emerge as we consider this work to be a first stepping-stone in improving planar scene representations via conditioning with extrinsic signals.
This includes the exploration of approaches to achieve this conditioning other than optimization guidance, and the application of scene refining on other representations.

\section*{Impact Statement}
This paper presents work that enhances the construction of high-quality neural representations. 
As such, the risks associated with our work parallel those of other neural rendering papers. 
This includes but is not limited to privacy and security concerns, as our method is trained on a dataset of publicly captured images, where privacy-sensitive information (e.g. human faces, license plate numbers) could be present. 
Hence, similarly to other neural rendering approaches, there is a risk that such data could end up in the trained model if the employed datasets are not properly filtered before use. 
Furthermore, as our work utilizes Stable Diffusion as prior, it inherits any problematic biases and limitations this model may have.

\section*{Acknowledgments}
This work was granted access to the HPC resources of IDRIS under the allocation 2023-AD011014261 made by GENCI.
We thank Loic Landrieu, Vicky Kalogeiton and Thibaut Issenhuth for inspiring discussions and valuable feedback. 

\bibliography{main}

\begin{thebibliography}{53}
\providecommand{\natexlab}[1]{#1}
\providecommand{\url}[1]{\texttt{#1}}
\expandafter\ifx\csname urlstyle\endcsname\relax
  \providecommand{\doi}[1]{doi: #1}\else
  \providecommand{\doi}{doi: \begingroup \urlstyle{rm}\Url}\fi

\bibitem[Aoi(2022)]{nerfw-public}
Al~Aoi.
\newblock nerf\_pl.
\newblock \url{https://github.com/kwea123/nerf_pl/tree/nerfw}, 2022.
\newblock Accessed: 2023-10-25.

\bibitem[Chan et~al.(2022)Chan, Lin, Chan, Nagano, Pan, De~Mello, Gallo, Guibas, Tremblay, Khamis, Karras, and Wetzstein]{eg3d}
Eric~R. Chan, Connor~Z. Lin, Matthew~A. Chan, Koki Nagano, Boxiao Pan, Shalini De~Mello, Orazio Gallo, Leonidas~J. Guibas, Jonathan Tremblay, Sameh Khamis, Tero Karras, and Gordon Wetzstein.
\newblock {Efficient Geometry-Aware 3D Generative Adversarial Networks}.
\newblock In \emph{Proceedings of the IEEE/CVF Conference on Computer Vision and Pattern Recognition (CVPR)}, pp.\  16123--16133, June 2022.

\bibitem[Chen et~al.(2022{\natexlab{a}})Chen, Xu, Geiger, Yu, and Su]{tensorf}
Anpei Chen, Zexiang Xu, Andreas Geiger, Jingyi Yu, and Hao Su.
\newblock {TensoRF: Tensorial Radiance Fields}.
\newblock In \emph{European Conference on Computer Vision (ECCV)}, 2022{\natexlab{a}}.

\bibitem[Chen et~al.(2024)Chen, Qin, Liu, Lu, and Li]{nerf-hugs}
Jiahao Chen, Yipeng Qin, Lingjie Liu, Jiangbo Lu, and Guanbin Li.
\newblock {NeRF-HuGS: Improved Neural Radiance Fields in Non-static Scenes Using Heuristics-Guided Segmentation}.
\newblock In \emph{Proceedings of the IEEE/CVF Conference on Computer Vision and Pattern Recognition (CVPR)}, pp.\  19436--19446, June 2024.

\bibitem[Chen et~al.(2022{\natexlab{b}})Chen, Zhang, Li, Chen, Feng, Wang, and Wang]{ha-nerf}
Xingyu Chen, Qi~Zhang, Xiaoyu Li, Yue Chen, Ying Feng, Xuan Wang, and Jue Wang.
\newblock {Hallucinated Neural Radiance Fields in the Wild}.
\newblock In \emph{Proceedings of the IEEE/CVF Conference on Computer Vision and Pattern Recognition (CVPR)}, pp.\  12943--12952, June 2022{\natexlab{b}}.

\bibitem[Deng et~al.(2023)Deng, Jiang, Qi, Yan, Zhou, Guibas, and Anguelov]{nerdi}
Congyue Deng, Chiyu~Max Jiang, Charles~R. Qi, Xinchen Yan, Yin Zhou, Leonidas Guibas, and Dragomir Anguelov.
\newblock Nerdi: Single-view nerf synthesis with language-guided diffusion as general image priors.
\newblock In \emph{Proceedings of the IEEE/CVF Conference on Computer Vision and Pattern Recognition (CVPR)}, pp.\  20637--20647, June 2023.

\bibitem[Fan et~al.(2023)Fan, Watkins, Du, Liu, Ryu, Boutilier, Abbeel, Ghavamzadeh, Lee, and Lee]{rl-ft-lora}
Ying Fan, Olivia Watkins, Yuqing Du, Hao Liu, Moonkyung Ryu, Craig Boutilier, Pieter Abbeel, Mohammad Ghavamzadeh, Kangwook Lee, and Kimin Lee.
\newblock Dpok: Reinforcement learning for fine-tuning text-to-image diffusion models.
\newblock \emph{arXiv preprint arXiv:2305.16381}, 2023.

\bibitem[Fridovich-Keil et~al.(2022)Fridovich-Keil, Yu, Tancik, Chen, Recht, and Kanazawa]{plenoxels}
Sara Fridovich-Keil, Alex Yu, Matthew Tancik, Qinhong Chen, Benjamin Recht, and Angjoo Kanazawa.
\newblock {Plenoxels: Radiance Fields Without Neural Networks}.
\newblock In \emph{Proceedings of the IEEE/CVF Conference on Computer Vision and Pattern Recognition (CVPR)}, pp.\  5501--5510, June 2022.

\bibitem[Fridovich-Keil et~al.(2023)Fridovich-Keil, Meanti, Warburg, Recht, and Kanazawa]{kplanes}
Sara Fridovich-Keil, Giacomo Meanti, Frederik~Rahb{\ae}k Warburg, Benjamin Recht, and Angjoo Kanazawa.
\newblock {K-Planes: Explicit Radiance Fields in Space, Time, and Appearance}.
\newblock In \emph{Proceedings of the IEEE/CVF Conference on Computer Vision and Pattern Recognition (CVPR)}, pp.\  12479--12488, June 2023.

\bibitem[Gao et~al.(2024)Gao, Holynski, Henzler, Brussee, Brualla, Srinivasan, Barron, and Poole]{cat3d}
Ruiqi Gao, Aleksander Holynski, Philipp Henzler, Arthur Brussee, Ricardo~Martin Brualla, Pratul~P. Srinivasan, Jonathan~T. Barron, and Ben Poole.
\newblock {CAT}3d: Create anything in 3d with multi-view diffusion models.
\newblock In \emph{The Thirty-eighth Annual Conference on Neural Information Processing Systems}, 2024.
\newblock URL \url{https://openreview.net/forum?id=TFZlFRl9Ks}.

\bibitem[Gimini(2016)]{bike}
Gianluca Gimini.
\newblock Velocipedia.
\newblock \url{https://www.gianlucagimini.it/portfolio-item/velocipedia/}, 2016.
\newblock Accessed: 2023-10-25.

\bibitem[Graikos et~al.(2022)Graikos, Malkin, Jojic, and Samaras]{diffusion-prior}
Alexandros Graikos, Nikolay Malkin, Nebojsa Jojic, and Dimitris Samaras.
\newblock {Diffusion Models as Plug-and-Play Priors}.
\newblock In S.~Koyejo, S.~Mohamed, A.~Agarwal, D.~Belgrave, K.~Cho, and A.~Oh (eds.), \emph{Advances in Neural Information Processing Systems}, volume~35, pp.\  14715--14728. Curran Associates, Inc., 2022.

\bibitem[Gu et~al.(2023)Gu, Trevithick, Lin, Susskind, Theobalt, Liu, and Ramamoorthi]{nerfdiff}
Jiatao Gu, Alex Trevithick, Kai-En Lin, Josh Susskind, Christian Theobalt, Lingjie Liu, and Ravi Ramamoorthi.
\newblock {NerfDiff: Single-image View Synthesis with NeRF-guided Distillation from 3D-aware Diffusion}.
\newblock In \emph{International Conference on Machine Learning}, 2023.

\bibitem[Gupta et~al.(2023)Gupta, Xiong, Nie, Jones, and O{\u{g}}uz]{3dgen}
Anchit Gupta, Wenhan Xiong, Yixin Nie, Ian Jones, and Barlas O{\u{g}}uz.
\newblock 3dgen: Triplane latent diffusion for textured mesh generation.
\newblock \emph{arXiv preprint arXiv:2303.05371}, 2023.

\bibitem[Hartley \& Zisserman(2004)Hartley and Zisserman]{sfm}
Richard Hartley and Andrew Zisserman.
\newblock \emph{Multiple View Geometry in Computer Vision}.
\newblock Cambridge University Press, 2 edition, 2004.
\newblock \doi{10.1017/CBO9780511811685}.

\bibitem[Ho et~al.(2020)Ho, Jain, and Abbeel]{ddpm}
Jonathan Ho, Ajay Jain, and Pieter Abbeel.
\newblock {Denoising Diffusion Probabilistic Models}.
\newblock In H.~Larochelle, M.~Ranzato, R.~Hadsell, M.F. Balcan, and H.~Lin (eds.), \emph{Advances in Neural Information Processing Systems}, volume~33, pp.\  6840--6851. Curran Associates, Inc., 2020.

\bibitem[Hu et~al.(2022)Hu, Shen, Wallis, Allen-Zhu, Li, Wang, Wang, and Chen]{lora}
Edward~J Hu, Yelong Shen, Phillip Wallis, Zeyuan Allen-Zhu, Yuanzhi Li, Shean Wang, Lu~Wang, and Weizhu Chen.
\newblock {LoRA: Low-Rank Adaptation of Large Language Models}.
\newblock In \emph{International Conference on Learning Representations}, 2022.

\bibitem[Jain et~al.(2021)Jain, Tancik, and Abbeel]{dietnerf}
Ajay Jain, Matthew Tancik, and Pieter Abbeel.
\newblock Putting nerf on a diet: Semantically consistent few-shot view synthesis.
\newblock In \emph{Proceedings of the IEEE/CVF International Conference on Computer Vision (ICCV)}, pp.\  5885--5894, October 2021.

\bibitem[Jin et~al.(2020)Jin, Mishkin, Mishchuk, Matas, Fua, Yi, and Trulls]{phototourism}
Yuhe Jin, Dmytro Mishkin, Anastasiia Mishchuk, Jiri Matas, Pascal Fua, Kwang~Moo Yi, and Eduard Trulls.
\newblock {Image Matching Across Wide Baselines: From Paper to Practice}.
\newblock \emph{International Journal of Computer Vision}, 129\penalty0 (2):\penalty0 517--547, oct 2020.
\newblock \doi{10.1007/s11263-020-01385-0}.

\bibitem[Kajiya \& Herzen(1984)Kajiya and Herzen]{volume-rendering}
James~T. Kajiya and Brian~Von Herzen.
\newblock {Ray tracing volume densities}.
\newblock \emph{Proceedings of the 11th annual conference on Computer graphics and interactive techniques}, 1984.

\bibitem[Kerbl et~al.(2023)Kerbl, Kopanas, Leimk{\"u}hler, and Drettakis]{gaussian-splatting}
Bernhard Kerbl, Georgios Kopanas, Thomas Leimk{\"u}hler, and George Drettakis.
\newblock {3D Gaussian Splatting for Real-Time Radiance Field Rendering}.
\newblock \emph{ACM Transactions on Graphics}, 42\penalty0 (4), July 2023.

\bibitem[Kulhanek et~al.(2024)Kulhanek, Peng, Kukelova, Pollefeys, and Sattler]{wildgaussians}
Jonas Kulhanek, Songyou Peng, Zuzana Kukelova, Marc Pollefeys, and Torsten Sattler.
\newblock {WildGaussians: 3D Gaussian Splatting In the Wild}.
\newblock In \emph{The Thirty-eighth Annual Conference on Neural Information Processing Systems}, 2024.

\bibitem[Lan et~al.(2024)Lan, Hong, Yang, Zhou, Meng, Dai, Pan, and Loy]{LN3Diff}
Yushi Lan, Fangzhou Hong, Shuai Yang, Shangchen Zhou, Xuyi Meng, Bo~Dai, Xingang Pan, and Chen~Change Loy.
\newblock {LN3Diff: Scalable Latent Neural Fields Diffusion for Speedy 3D Generation}.
\newblock In \emph{ECCV}, 2024.

\bibitem[Lee et~al.(2023)Lee, Hunter, and Ruiz]{lee2023platypus}
Ariel~N Lee, Cole~J Hunter, and Nataniel Ruiz.
\newblock Platypus: Quick, cheap, and powerful refinement of llms.
\newblock \emph{arXiv preprint arXiv:2308.07317}, 2023.

\bibitem[Liu et~al.(2023)Liu, Wu, Van~Hoorick, Tokmakov, Zakharov, and Vondrick]{zero-1-to-3}
Ruoshi Liu, Rundi Wu, Basile Van~Hoorick, Pavel Tokmakov, Sergey Zakharov, and Carl Vondrick.
\newblock {Zero-1-to-3: Zero-shot One Image to 3D Object}.
\newblock In \emph{Proceedings of the IEEE/CVF International Conference on Computer Vision (ICCV)}, pp.\  9298--9309, October 2023.

\bibitem[Liu et~al.(2024)Liu, Guo, Luo, Sun, Yin, and Zhang]{pi3d}
Ying-Tian Liu, Yuan-Chen Guo, Guan Luo, Heyi Sun, Wei Yin, and Song-Hai Zhang.
\newblock {PI3D: Efficient Text-to-3D Generation with Pseudo-Image Diffusion}.
\newblock In \emph{Proceedings of the IEEE/CVF Conference on Computer Vision and Pattern Recognition (CVPR)}, pp.\  19915--19924, June 2024.

\bibitem[Martin-Brualla et~al.(2021)Martin-Brualla, Radwan, Sajjadi, Barron, Dosovitskiy, and Duckworth]{nerf-w}
Ricardo Martin-Brualla, Noha Radwan, Mehdi S.~M. Sajjadi, Jonathan~T. Barron, Alexey Dosovitskiy, and Daniel Duckworth.
\newblock {NeRF in the Wild: Neural Radiance Fields for Unconstrained Photo Collections}.
\newblock In \emph{Proceedings of the IEEE/CVF Conference on Computer Vision and Pattern Recognition (CVPR)}, pp.\  7210--7219, June 2021.

\bibitem[Melas-Kyriazi et~al.(2023)Melas-Kyriazi, Laina, Rupprecht, and Vedaldi]{realfusion}
Luke Melas-Kyriazi, Iro Laina, Christian Rupprecht, and Andrea Vedaldi.
\newblock {RealFusion: 360deg Reconstruction of Any Object From a Single Image}.
\newblock In \emph{Proceedings of the IEEE/CVF Conference on Computer Vision and Pattern Recognition (CVPR)}, pp.\  8446--8455, June 2023.

\bibitem[Metzer et~al.(2023)Metzer, Richardson, Patashnik, Giryes, and Cohen-Or]{latentnerf}
Gal Metzer, Elad Richardson, Or~Patashnik, Raja Giryes, and Daniel Cohen-Or.
\newblock {Latent-NeRF for Shape-Guided Generation of 3D Shapes and Textures}.
\newblock In \emph{Proceedings of the IEEE/CVF Conference on Computer Vision and Pattern Recognition (CVPR)}, pp.\  12663--12673, June 2023.

\bibitem[Mildenhall et~al.(2020)Mildenhall, Srinivasan, Tancik, Barron, Ramamoorthi, and Ng]{nerf}
Ben Mildenhall, Pratul~P. Srinivasan, Matthew Tancik, Jonathan~T. Barron, Ravi Ramamoorthi, and Ren Ng.
\newblock {NeRF: Representing Scenes as Neural Radiance Fields for View Synthesis}.
\newblock In \emph{ECCV}, 2020.

\bibitem[M\"uller(2021)]{tiny-cuda-nn}
Thomas M\"uller.
\newblock {tiny-cuda-nn}, 4 2021.
\newblock URL \url{https://github.com/NVlabs/tiny-cuda-nn}.

\bibitem[M\"uller et~al.(2022)M\"uller, Evans, Schied, and Keller]{instantngp}
Thomas M\"uller, Alex Evans, Christoph Schied, and Alexander Keller.
\newblock {Instant Neural Graphics Primitives with a Multiresolution Hash Encoding}.
\newblock \emph{ACM Trans. Graph.}, 41\penalty0 (4):\penalty0 102:1--102:15, July 2022.
\newblock \doi{10.1145/3528223.3530127}.

\bibitem[Paszke et~al.(2019)Paszke, Gross, Massa, Lerer, Bradbury, Chanan, Killeen, Lin, Gimelshein, Antiga, Desmaison, Kopf, Yang, DeVito, Raison, Tejani, Chilamkurthy, Steiner, Fang, Bai, and Chintala]{pytorch}
Adam Paszke, Sam Gross, Francisco Massa, Adam Lerer, James Bradbury, Gregory Chanan, Trevor Killeen, Zeming Lin, Natalia Gimelshein, Luca Antiga, Alban Desmaison, Andreas Kopf, Edward Yang, Zachary DeVito, Martin Raison, Alykhan Tejani, Sasank Chilamkurthy, Benoit Steiner, Lu~Fang, Junjie Bai, and Soumith Chintala.
\newblock {PyTorch: An Imperative Style, High-Performance Deep Learning Library}.
\newblock In H.~Wallach, H.~Larochelle, A.~Beygelzimer, F.~d\textquotesingle Alch\'{e}-Buc, E.~Fox, and R.~Garnett (eds.), \emph{Advances in Neural Information Processing Systems}, volume~32. Curran Associates, Inc., 2019.

\bibitem[Poole et~al.(2023)Poole, Jain, Barron, and Mildenhall]{dreamfusion}
Ben Poole, Ajay Jain, Jonathan~T. Barron, and Ben Mildenhall.
\newblock {DreamFusion: Text-to-3D using 2D Diffusion}.
\newblock In \emph{The Eleventh International Conference on Learning Representations}, 2023.

\bibitem[Rombach et~al.(2022)Rombach, Blattmann, Lorenz, Esser, and Ommer]{stable-diffusion}
Robin Rombach, Andreas Blattmann, Dominik Lorenz, Patrick Esser, and Bj\"orn Ommer.
\newblock {High-Resolution Image Synthesis With Latent Diffusion Models}.
\newblock In \emph{Proceedings of the IEEE/CVF Conference on Computer Vision and Pattern Recognition (CVPR)}, pp.\  10684--10695, June 2022.

\bibitem[Sch\"{o}nberger \& Frahm(2016)Sch\"{o}nberger and Frahm]{colmap}
Johannes~Lutz Sch\"{o}nberger and Jan-Michael Frahm.
\newblock Structure-from-motion revisited.
\newblock In \emph{Conference on Computer Vision and Pattern Recognition (CVPR)}, 2016.

\bibitem[Shi et~al.(2023)Shi, Wang, Ye, Long, Li, and Yang]{mvdream}
Yichun Shi, Peng Wang, Jianglong Ye, Mai Long, Kejie Li, and Xiao Yang.
\newblock Mvdream: Multi-view diffusion for 3d generation, 2023.

\bibitem[Shue et~al.(2023)Shue, Chan, Po, Ankner, Wu, and Wetzstein]{3D-nf-gen-triplane-diffusion}
J.~Ryan Shue, Eric~Ryan Chan, Ryan Po, Zachary Ankner, Jiajun Wu, and Gordon Wetzstein.
\newblock {3D Neural Field Generation Using Triplane Diffusion}.
\newblock In \emph{Proceedings of the IEEE/CVF Conference on Computer Vision and Pattern Recognition (CVPR)}, pp.\  20875--20886, June 2023.

\bibitem[Shum et~al.(2008)Shum, Chan, and Kang]{image-based-rendering}
H.Y. Shum, S.C. Chan, and S.B. Kang.
\newblock \emph{Image-Based Rendering}.
\newblock Springer US, 2008.
\newblock ISBN 9780387326689.

\bibitem[Tancik et~al.(2023)Tancik, Weber, Ng, Li, Yi, Kerr, Wang, Kristoffersen, Austin, Salahi, Ahuja, McAllister, and Kanazawa]{nerfstudio}
Matthew Tancik, Ethan Weber, Evonne Ng, Ruilong Li, Brent Yi, Justin Kerr, Terrance Wang, Alexander Kristoffersen, Jake Austin, Kamyar Salahi, Abhik Ahuja, David McAllister, and Angjoo Kanazawa.
\newblock {Nerfstudio: A Modular Framework for Neural Radiance Field Development}.
\newblock In \emph{ACM SIGGRAPH 2023 Conference Proceedings}, SIGGRAPH '23, 2023.

\bibitem[Tewari et~al.(2020)Tewari, Fried, Thies, Sitzmann, Lombardi, Sunkavalli, Martin-Brualla, Simon, Saragih, Nießner, Pandey, Fanello, Wetzstein, Zhu, Theobalt, Agrawala, Shechtman, Goldman, and Zollhöfer]{sota-nr}
Ayush Tewari, Ohad Fried, Justus Thies, Vincent Sitzmann, Stephen Lombardi, Kalyan Sunkavalli, Ricardo Martin-Brualla, Tomas Simon, Jason Saragih, Matthias Nießner, Rohit Pandey, Sean Fanello, Gordon Wetzstein, Jun-Yan Zhu, Christian Theobalt, Maneesh Agrawala, Eli Shechtman, Dan~B. Goldman, and Michael Zollhöfer.
\newblock {State of the Art on Neural Rendering}.
\newblock \emph{Computer Graphics Forum (Eurographics '20): State of the Art Reports}, 39\penalty0 (2):\penalty0 701 -- 727, May 2020.

\bibitem[von Platen et~al.(2022)von Platen, Patil, Lozhkov, Cuenca, Lambert, Rasul, Davaadorj, and Wolf]{diffusers}
Patrick von Platen, Suraj Patil, Anton Lozhkov, Pedro Cuenca, Nathan Lambert, Kashif Rasul, Mishig Davaadorj, and Thomas Wolf.
\newblock Diffusers: State-of-the-art diffusion models, 2022.
\newblock Accessed: 2023-10-25.

\bibitem[Wang et~al.(2023{\natexlab{a}})Wang, Yue, Zhou, Chan, and Loy]{sd-for-super-resolution}
Jianyi Wang, Zongsheng Yue, Shangchen Zhou, Kelvin~CK Chan, and Chen~Change Loy.
\newblock {Exploiting Diffusion Prior for Real-World Image Super-Resolution}.
\newblock In \emph{arXiv preprint arXiv:2305.07015}, 2023{\natexlab{a}}.

\bibitem[Wang et~al.(2024)Wang, Leroy, Cabon, Chidlovskii, and Revaud]{dust3r}
Shuzhe Wang, Vincent Leroy, Yohann Cabon, Boris Chidlovskii, and Jerome Revaud.
\newblock Dust3r: Geometric 3d vision made easy.
\newblock In \emph{Proceedings of the IEEE/CVF Conference on Computer Vision and Pattern Recognition (CVPR)}, pp.\  20697--20709, June 2024.

\bibitem[Wang et~al.(2023{\natexlab{b}})Wang, Zhang, Zhang, Gu, Bao, Baltrusaitis, Shen, Chen, Wen, Chen, and Guo]{rodin}
Tengfei Wang, Bo~Zhang, Ting Zhang, Shuyang Gu, Jianmin Bao, Tadas Baltrusaitis, Jingjing Shen, Dong Chen, Fang Wen, Qifeng Chen, and Baining Guo.
\newblock {RODIN: A Generative Model for Sculpting 3D Digital Avatars Using Diffusion}.
\newblock In \emph{Proceedings of the IEEE/CVF Conference on Computer Vision and Pattern Recognition (CVPR)}, pp.\  4563--4573, June 2023{\natexlab{b}}.

\bibitem[Watson et~al.(2023)Watson, Chan, Brualla, Ho, Tagliasacchi, and Norouzi]{nvs-diffusion-models}
Daniel Watson, William Chan, Ricardo~Martin Brualla, Jonathan Ho, Andrea Tagliasacchi, and Mohammad Norouzi.
\newblock {Novel View Synthesis with Diffusion Models}.
\newblock In \emph{The Eleventh International Conference on Learning Representations}, 2023.

\bibitem[Wynn \& Turmukhambetov(2023)Wynn and Turmukhambetov]{diffusionerf}
Jamie Wynn and Daniyar Turmukhambetov.
\newblock {DiffusioNeRF: Regularizing Neural Radiance Fields With Denoising Diffusion Models}.
\newblock In \emph{Proceedings of the IEEE/CVF Conference on Computer Vision and Pattern Recognition (CVPR)}, pp.\  4180--4189, June 2023.

\bibitem[Xu et~al.(2024)Xu, Kerr, and Kanazawa]{splatfacto-w}
Congrong Xu, Justin Kerr, and Angjoo Kanazawa.
\newblock {Splatfacto-w: A nerfstudio implementation of gaussian splatting for unconstrained photo collections}.
\newblock \emph{arXiv preprint arXiv:2407.12306}, 2024.

\bibitem[Yang et~al.(2023)Yang, Zhang, Huang, Zhang, and Tan]{cr-nerf}
Yifan Yang, Shuhai Zhang, Zixiong Huang, Yubing Zhang, and Mingkui Tan.
\newblock {Cross-Ray Neural Radiance Fields for Novel-View Synthesis from Unconstrained Image Collections}.
\newblock In \emph{Proceedings of the IEEE/CVF International Conference on Computer Vision (ICCV)}, pp.\  15901--15911, October 2023.

\bibitem[Yu et~al.(2021)Yu, Ye, Tancik, and Kanazawa]{pixelnerf}
Alex Yu, Vickie Ye, Matthew Tancik, and Angjoo Kanazawa.
\newblock pixelnerf: Neural radiance fields from one or few images.
\newblock In \emph{Proceedings of the IEEE/CVF Conference on Computer Vision and Pattern Recognition (CVPR)}, pp.\  4578--4587, June 2021.

\bibitem[Zeng \& Lee(2023)Zeng and Lee]{expressive-lora}
Yuchen Zeng and Kangwook Lee.
\newblock The expressive power of low-rank adaptation.
\newblock \emph{arXiv preprint arXiv:2310.17513}, 2023.

\bibitem[Zhang et~al.(2025)Zhang, Wang, Wang, Li, Qin, and Wang]{gs-w}
Dongbin Zhang, Chuming Wang, Weitao Wang, Peihao Li, Minghan Qin, and Haoqian Wang.
\newblock {Gaussian in the wild: 3d gaussian splatting for unconstrained image collections}.
\newblock In \emph{European Conference on Computer Vision}, pp.\  341--359. Springer, 2025.

\bibitem[Zhang et~al.(2018)Zhang, Isola, Efros, Shechtman, and Wang]{lpips}
Richard Zhang, Phillip Isola, Alexei~A. Efros, Eli Shechtman, and Oliver Wang.
\newblock {The Unreasonable Effectiveness of Deep Features as a Perceptual Metric}.
\newblock In \emph{Proceedings of the IEEE Conference on Computer Vision and Pattern Recognition (CVPR)}, June 2018.

\end{thebibliography}
\bibliographystyle{tmlr}

\newpage
\appendix
\section{Experimental Details}
\label{x:exp-details}
\subsection{Datasets}
\label{x:dataset-details}
\paragraph{Synthetic dataset.} For synthetic renderings, we adopt the \emph{Real Synthetic $\mathit{360^{\circ}}$} dataset from NeRF \citep{nerf}. This dataset consists of eight path-traced scenes containing objects exhibiting complicated geometry and realistic non-Lambertian materials. Each image is coupled with its corresponding camera parameters. Consistently with prior work, 100 images are used for training each scene and 200 images are used for testing. All images are at $800 \times 800$ pixels.

\paragraph{In-the-wild dataset. } For in-the-wild renderings, we adopt the Phototourism dataset \citep{phototourism} which is commonly used for in-the-wild tasks. This dataset consists of multitudes of images of touristic landmarks gathered from the internet. Thus, these images are naturally plagued by visual discrepancies, notably illumination variation and transient occluders. Camera parameters are estimated using COLMAP \citep{colmap}. All images are normalized to [0,1]. We adopt three scenes from Phototurism: \emph{Brandenburg Gate} (1363 images), \emph{Sacré Coeur} (1179 images), and \emph{Trevi Fountain} (3191 images). Testing is done on a standard set that is free of transient occluders.

\subsection{Frameworks}
\label{x:frameworks-details}
We make use of multiple frameworks to implement our method. 
Our Python source code (tested on version 3.7.16), based on PyTorch \citep{pytorch} (tested on version 1.13.1) and CUDA (tested on version 11.6), is publicly available as open source. 
We also utilize Diffusers \citep{diffusers} and Stable Diffusion (tested on version 1-5, main revision). 
K-Planes also adopt the \emph{tinycudann} framework \citep{tiny-cuda-nn}. 
We run all experiments on a single NVIDIA A100 GPU.

\subsection{Hyperparameters}
\label{x:hyperparameter-details}
A summary of our hyperparameters for synthetic as well as in-the-wild scenes can be found in \cref{table:hyp}.

\section{Feature Planes Inspection}
\label{x:inspection}
\begin{figure}[H]
    \centering
    \subfloat{\includegraphics[width=0.36\linewidth]{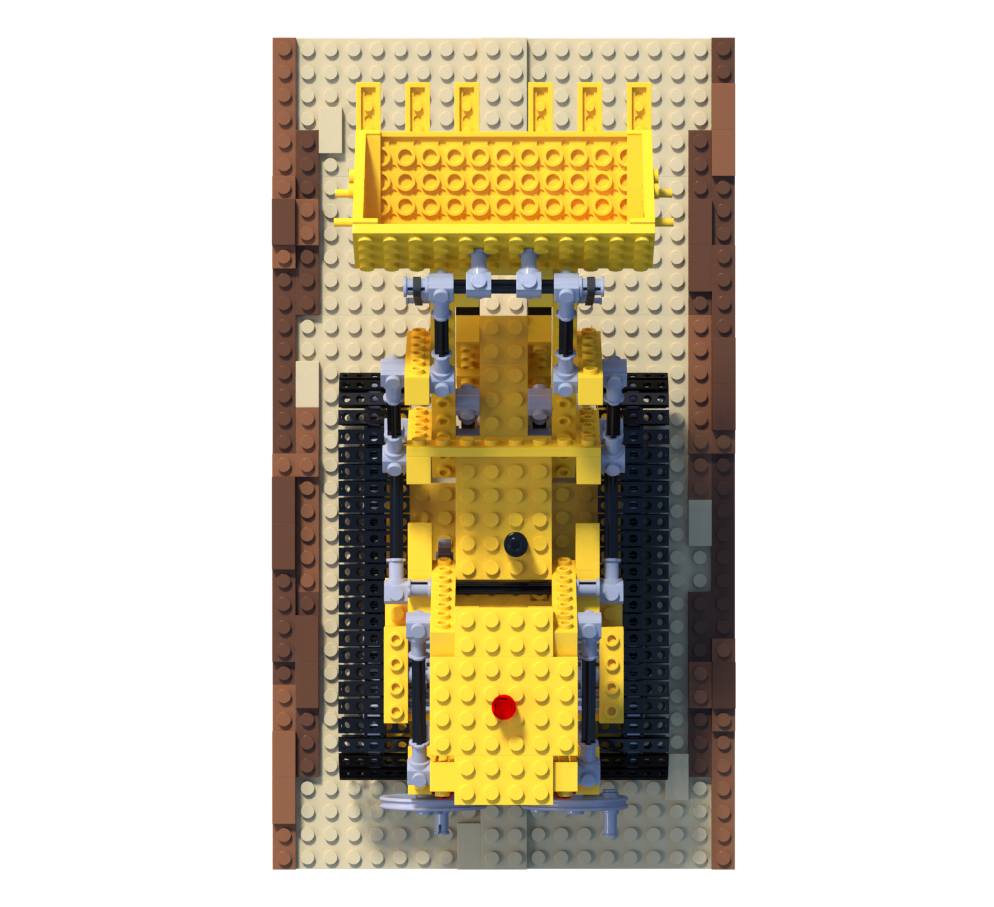}}
    \\
    \begin{tabular}{ccc}
        \toprule
        \subfloat{\includegraphics[width=0.25\linewidth]{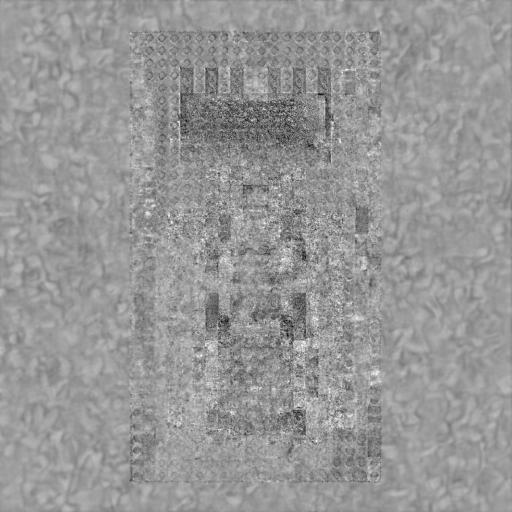}} & \hspace{-10pt}
        \subfloat{\includegraphics[width=0.25\linewidth]{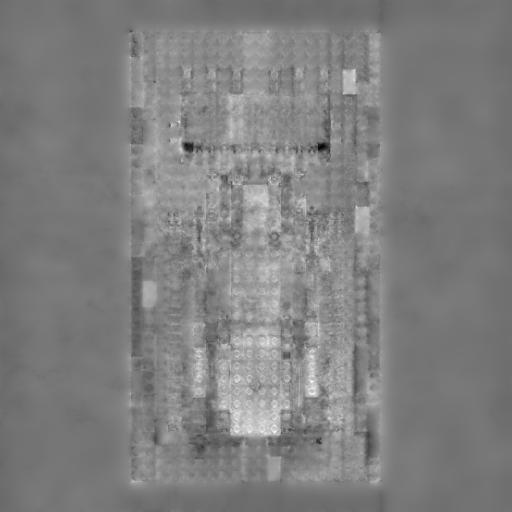}} & \hspace{-10pt}
        \subfloat{\includegraphics[width=0.25\linewidth]{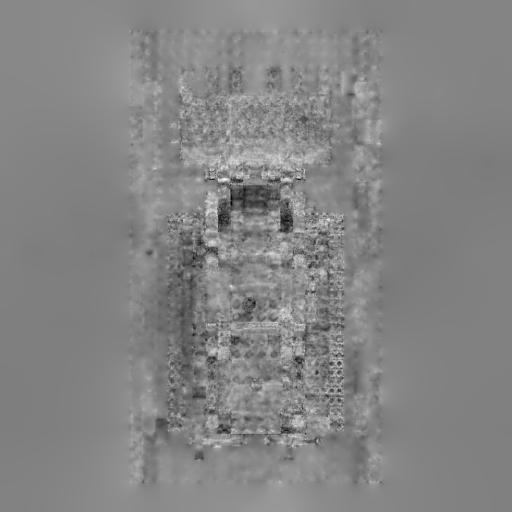}}
    \end{tabular}
    \\
    \vspace{-10pt}
    \begin{tabular}{ccc}
        \subfloat{\includegraphics[width=0.25\linewidth]{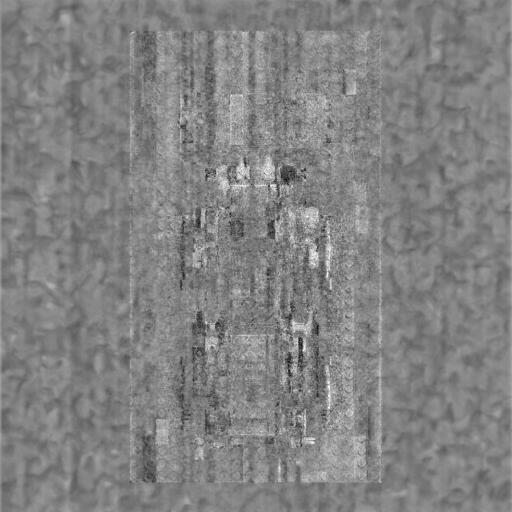}} & \hspace{-10pt}
        \subfloat{\includegraphics[width=0.25\linewidth]{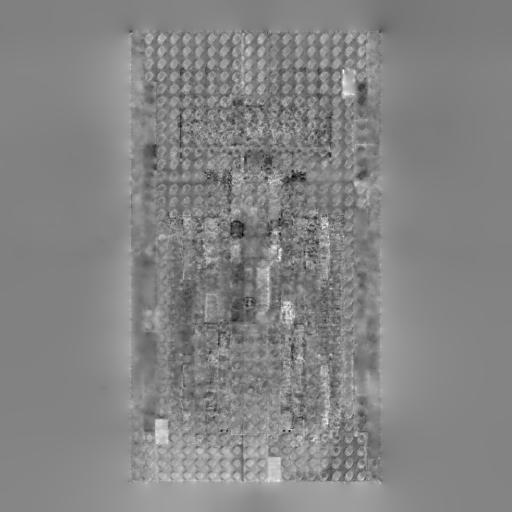}} & \hspace{-10pt}
        \subfloat{\includegraphics[width=0.25\linewidth]{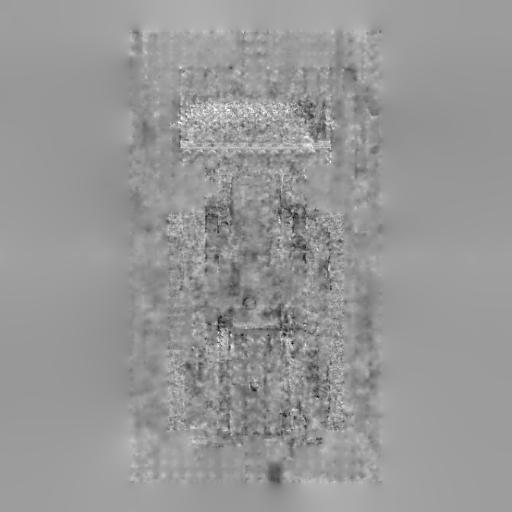}}
    \end{tabular}
    \\
    \vspace{-10pt}
    \begin{tabular}{ccc}
        \setcounter{subfigure}{0}
        \subfloat[RefinedFields\\(In-Process)]{
            \includegraphics[width=0.25\linewidth]{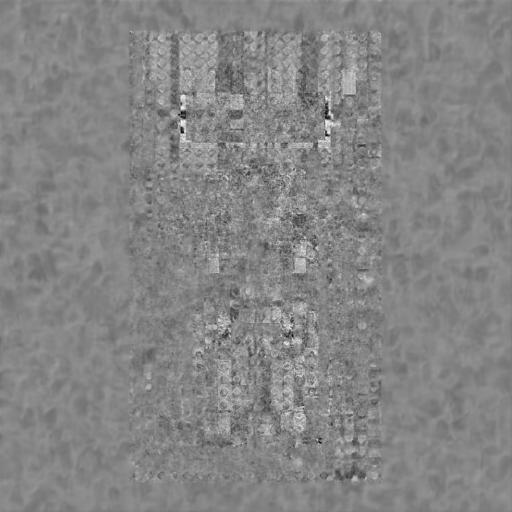}
            \label{x:insp-in-process-0}
        } & \hspace{-20pt}
        \subfloat[RefinedFields\\(Final)]{
            \includegraphics[width=0.25\linewidth]{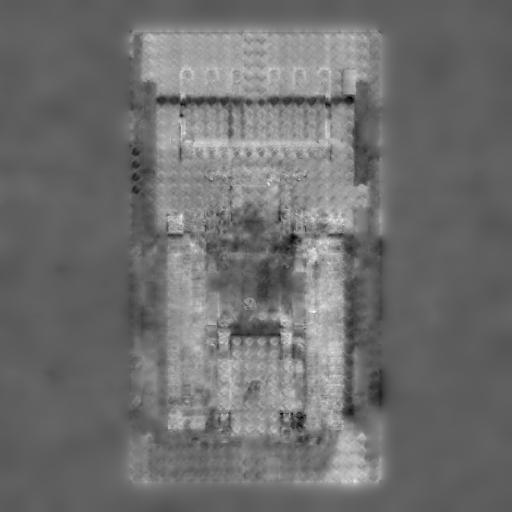}
            \label{x:insp-refinedfields-0}
        } & \hspace{-20pt}
        \subfloat[K-Planes]{
            \includegraphics[width=0.25\linewidth]{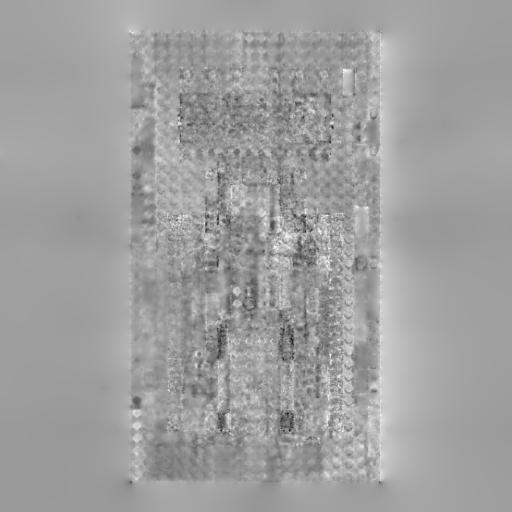}
            \label{x:insp-kplanes-0}
        }
    \end{tabular}
    \caption{\textbf{Feature planes inspection.} Visualization of the $(xy)$ K-Planes feature planes during the RefinedFields optimization process (\ref{x:insp-in-process-0}), at the end of the RefinedFields optimization (\ref{x:insp-refinedfields-0}), and a comparison with vanilla K-Planes-SS (\ref{x:insp-kplanes-0}). Feature planes within the $(xy)$ K-Planes are picked randomly.}
    \label{fig:inspection-0}
\end{figure}

\begin{figure}[H]
    \centering
    \subfloat{\includegraphics[width=0.3\linewidth]{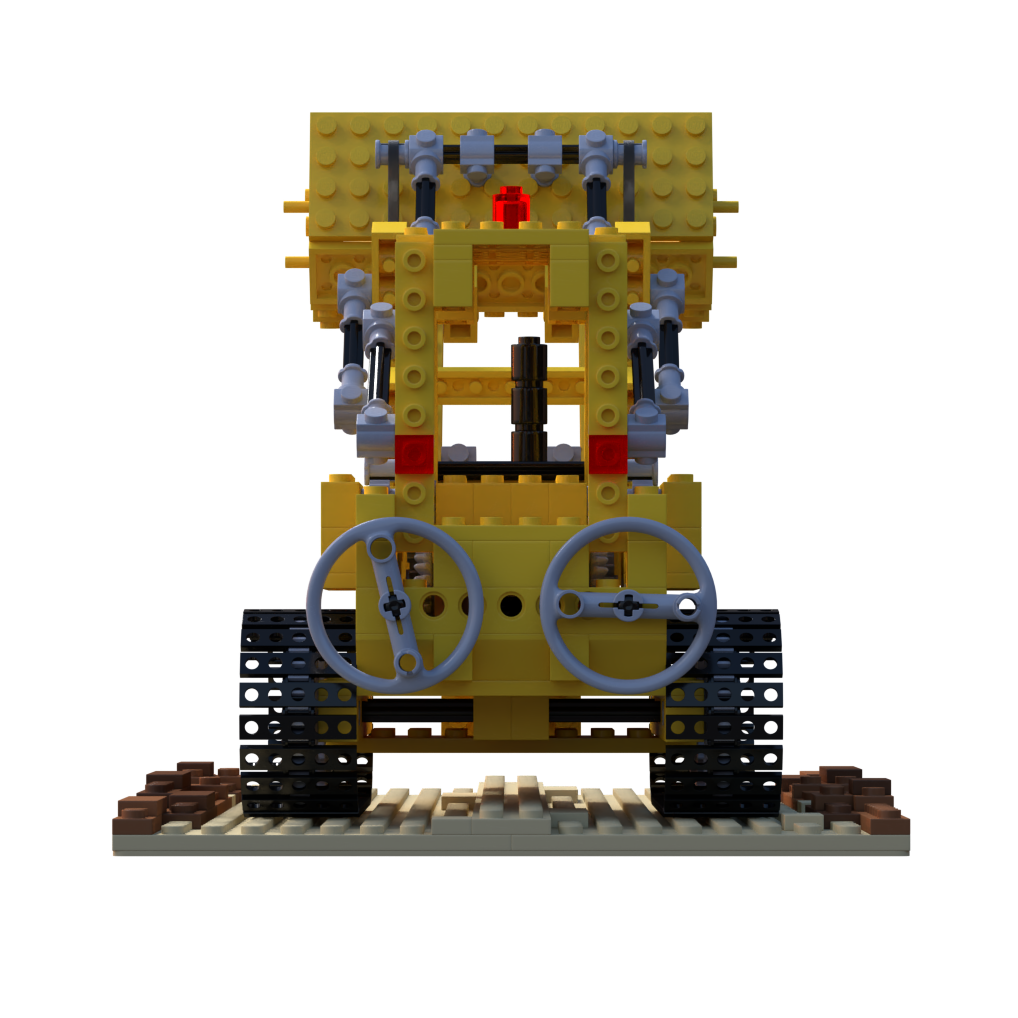}}
    \\
    \begin{tabular}{ccc}
        \toprule
        \subfloat{\includegraphics[width=0.25\linewidth,angle=180,origin=c]{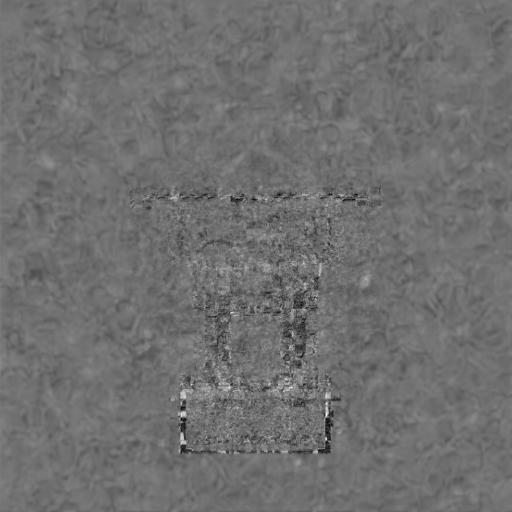}} & \hspace{-10pt}
        \subfloat{\includegraphics[width=0.25\linewidth,angle=180,origin=c]{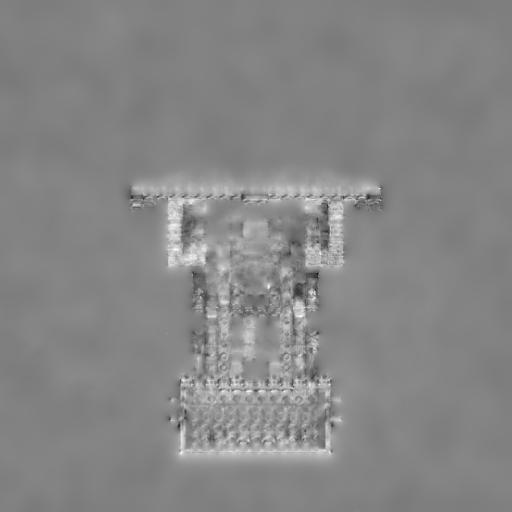}} & \hspace{-10pt}
        \subfloat{\includegraphics[width=0.25\linewidth,angle=180,origin=c]{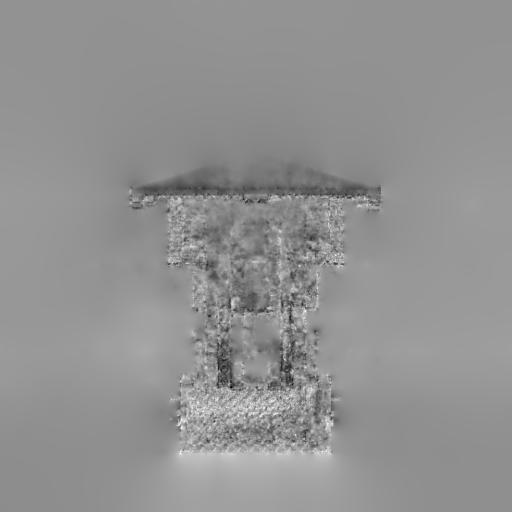}}
    \end{tabular}
    \\
    \vspace{-10pt}
    \begin{tabular}{ccc}
        \subfloat{\includegraphics[width=0.25\linewidth,angle=180,origin=c]{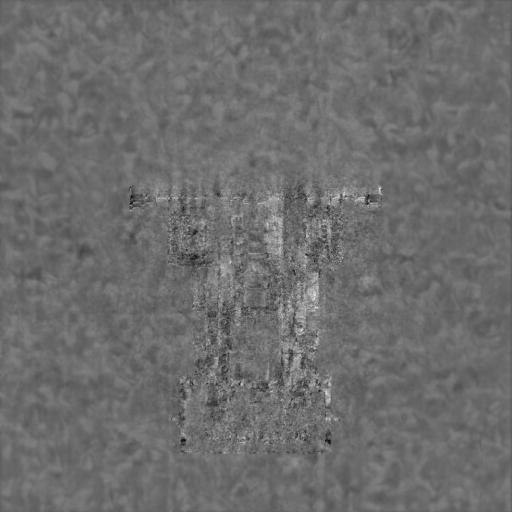}} & \hspace{-10pt}
        \subfloat{\includegraphics[width=0.25\linewidth,angle=180,origin=c]{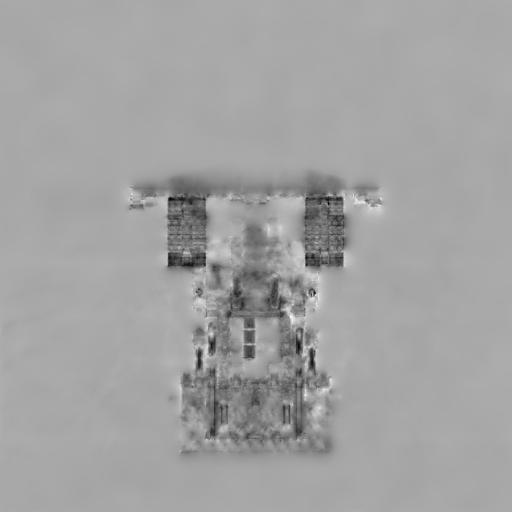}} & \hspace{-10pt}
        \subfloat{\includegraphics[width=0.25\linewidth,angle=180,origin=c]{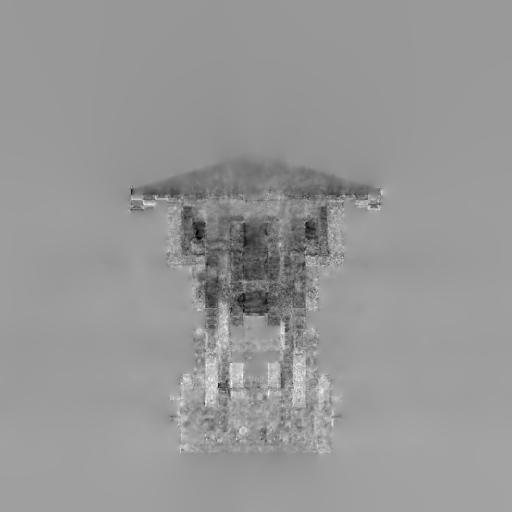}}
    \end{tabular}
    \\
    \vspace{-10pt}
    \begin{tabular}{ccc}
        \setcounter{subfigure}{0}
        \subfloat[RefinedFields\\(In-Process)]{
            \includegraphics[width=0.25\linewidth,angle=180,origin=c]{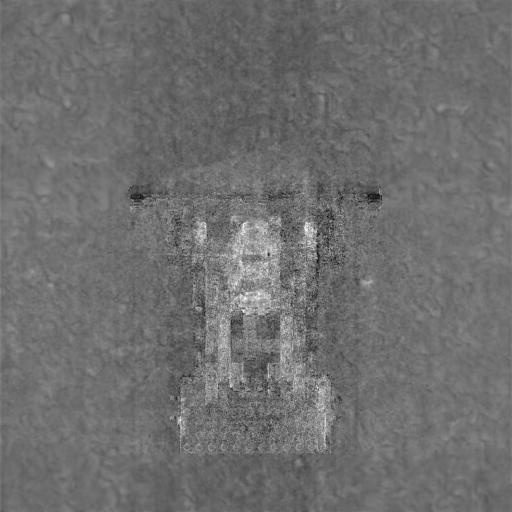}
            \label{x:insp-in-process-1}
        } & \hspace{-20pt}
        \subfloat[RefinedFields\\(Final)]{
            \includegraphics[width=0.25\linewidth,angle=180,origin=c]{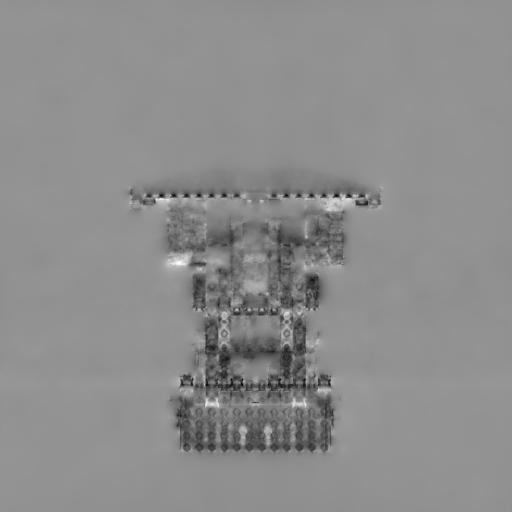}
            \label{x:insp-refinedfields-1}
        } & \hspace{-20pt}
        \subfloat[K-Planes]{
            \includegraphics[width=0.25\linewidth,angle=180,origin=c]{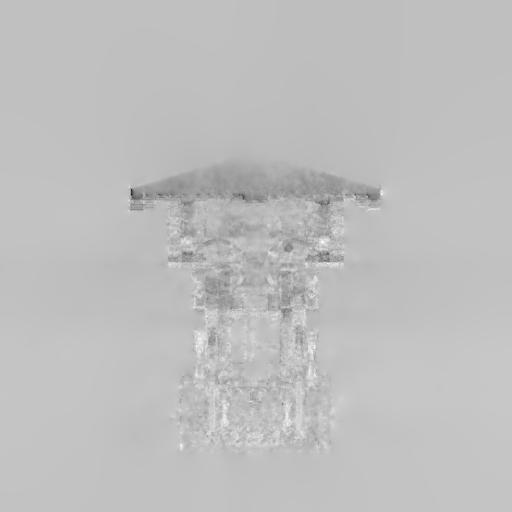}
            \label{x:insp-kplanes-1}
        }
    \end{tabular}
    \caption{\textbf{Feature planes inspection.} Visualization of the $(xz)$ K-Planes feature planes during the RefinedFields optimization process (\ref{x:insp-in-process-1}), at the end of the RefinedFields optimization (\ref{x:insp-refinedfields-1}), and a comparison with vanilla K-Planes-SS (\ref{x:insp-kplanes-1}). Feature planes within the $(xz)$ K-Planes are picked randomly.}
    \label{fig:inspection-1}
\end{figure}

\begin{figure}[H]
    \centering
    \subfloat{\includegraphics[width=0.35\linewidth]{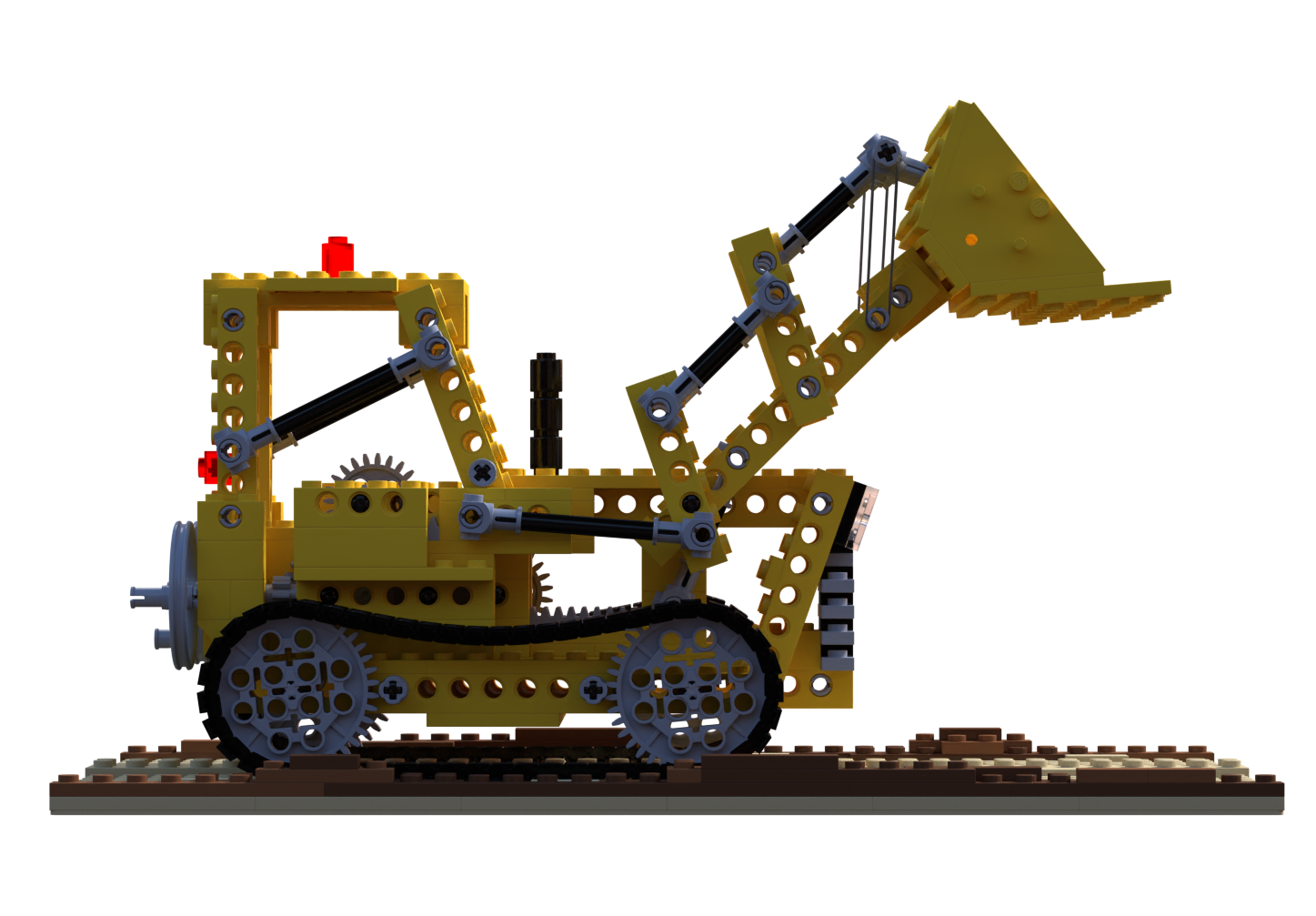}}
    \\
    \begin{tabular}{ccc}
        \toprule
        \subfloat{\includegraphics[width=0.25\linewidth,angle=180,origin=c]{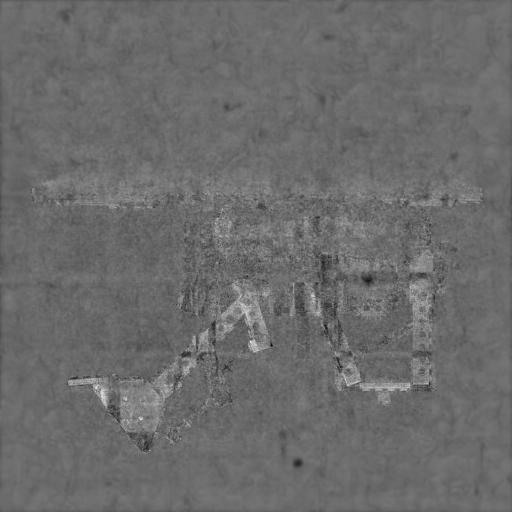}} & \hspace{-10pt}
        \subfloat{\includegraphics[width=0.25\linewidth,angle=180,origin=c]{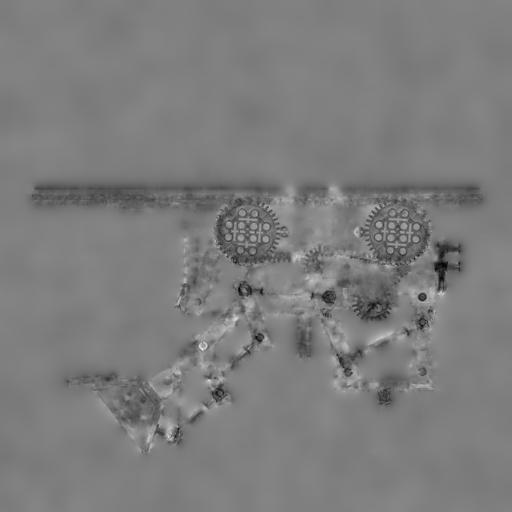}} & \hspace{-10pt}
        \subfloat{\includegraphics[width=0.25\linewidth,angle=180,origin=c]{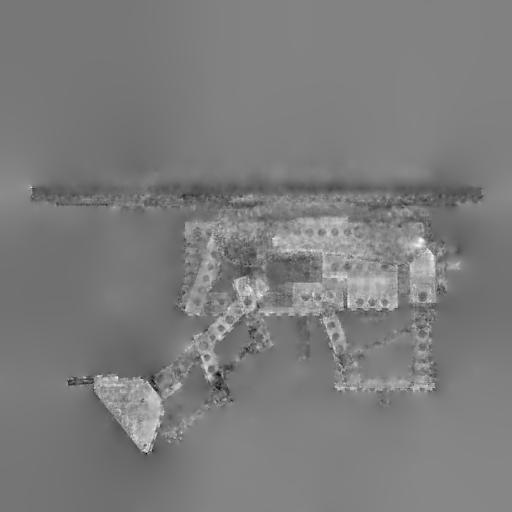}}
    \end{tabular}
    \\
    \vspace{-10pt}
    \begin{tabular}{ccc}
        \subfloat{\includegraphics[width=0.25\linewidth,angle=180,origin=c]{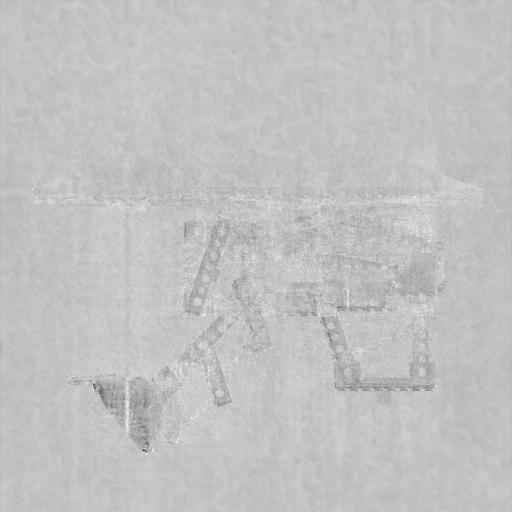}} & \hspace{-10pt}
        \subfloat{\includegraphics[width=0.25\linewidth,angle=180,origin=c]{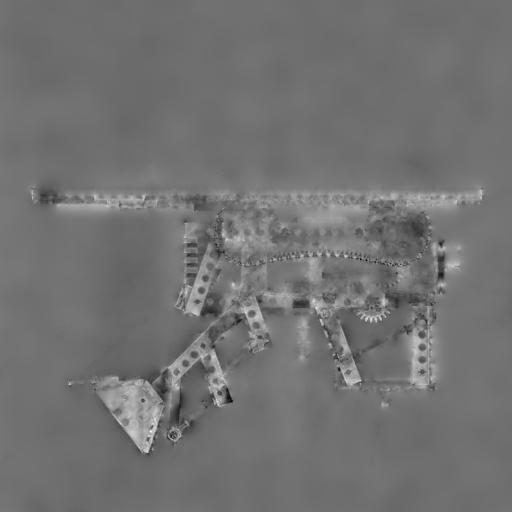}} & \hspace{-10pt}
        \subfloat{\includegraphics[width=0.25\linewidth,angle=180,origin=c]{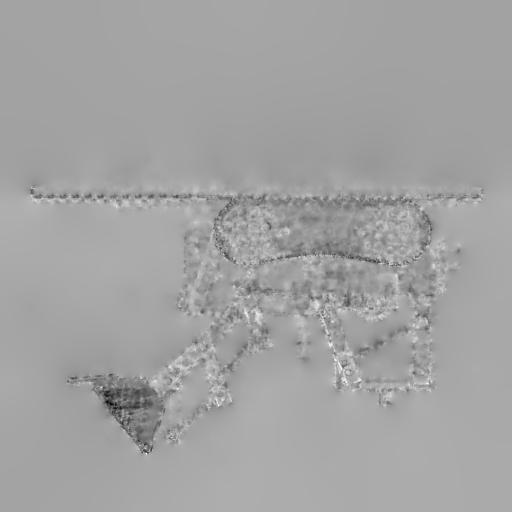}}
    \end{tabular}
    \\
    \vspace{-10pt}
    \begin{tabular}{ccc}
        \setcounter{subfigure}{0}
        \subfloat[RefinedFields\\(In-Process)]{
            \includegraphics[width=0.25\linewidth,angle=180,origin=c]{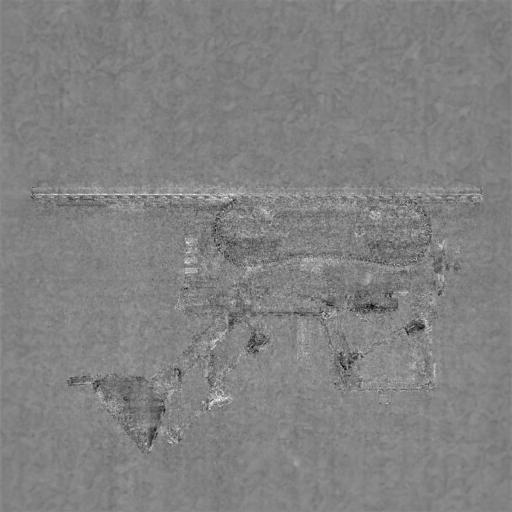}
            \label{x:insp-in-process-2}
        } & \hspace{-20pt}
        \subfloat[RefinedFields\\(Final)]{
            \includegraphics[width=0.25\linewidth,angle=180,origin=c]{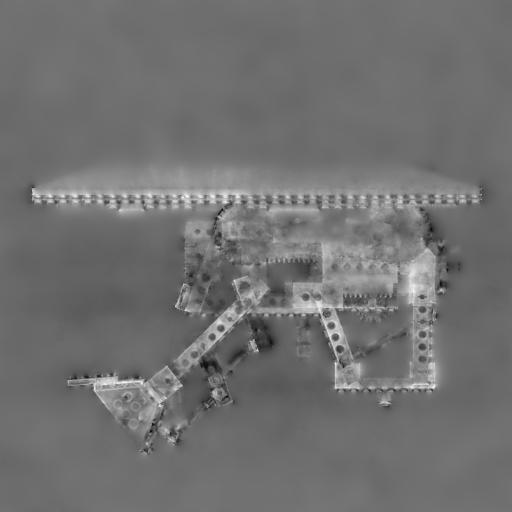}
            \label{x:insp-refinedfields-2}
        } & \hspace{-20pt}
        \subfloat[K-Planes]{
            \includegraphics[width=0.25\linewidth,angle=180,origin=c]{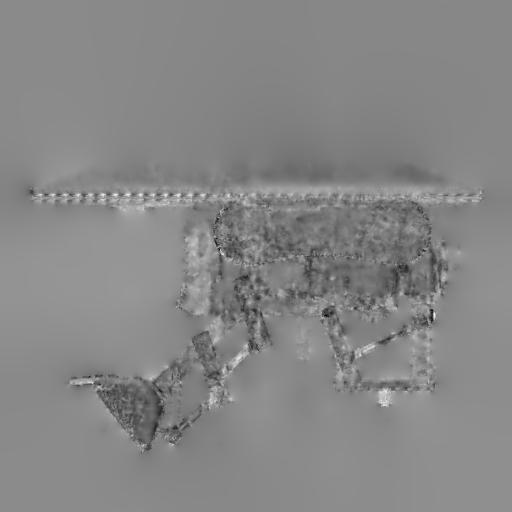}
            \label{x:insp-kplanes-2}
        }
    \end{tabular}
    \caption{\textbf{Feature planes inspection.} Visualization of the $(yz)$ K-Planes feature planes during the RefinedFields optimization process (\ref{x:insp-in-process-2}), at the end of the RefinedFields optimization (\ref{x:insp-refinedfields-2}), and a comparison with vanilla K-Planes-SS (\ref{x:insp-kplanes-2}). Feature planes within the $(yz)$ K-Planes are picked randomly.}
    \label{fig:inspection-2}
\end{figure}

In this section, we present a visual inspection of K-Planes feature planes within different contexts. \cref{fig:inspection-0,fig:inspection-1,fig:inspection-2} each correspond to one out of the three orthogonal planes. Each element in \cref{fig:inspection-0,fig:inspection-1,fig:inspection-2} presents a single feature plane, picked randomly from the $C$ feature planes. \cref{x:insp-in-process-0,x:insp-in-process-1,x:insp-in-process-2} represent the state of the planes at an intermediate stage of the RefinedFields optimization process. \cref{x:insp-refinedfields-0,x:insp-refinedfields-1,x:insp-refinedfields-2} represent the state of the planes at the end of the RefinedFields optimization. \cref{x:insp-kplanes-0,x:insp-kplanes-1,x:insp-kplanes-2} represent the planes at the end of the K-Planes-SS optimization (no refining is done in this case).

Two noteworthy observations emerge. First, as seen in \cref{fig:inspection-0,fig:inspection-1,fig:inspection-2}, K-Planes feature planes are very similar in structure to real images. In fact, these feature planes depict orthogonal projections of the scene onto the planes. These findings are especially compelling, as they justify the appropriate choice of Stable Diffusion as the pre-trained prior for the refining stage, and provide insight onto the quantitative and qualitative results showcased by our method. Second, a comparison between columns [\labelcref{x:insp-refinedfields-0,x:insp-refinedfields-1,x:insp-refinedfields-2}] and [\labelcref{x:insp-kplanes-0,x:insp-kplanes-1,x:insp-kplanes-2}] highlights the impact scene refining has on the feature planes themselves, as planes \labelcref{x:insp-refinedfields-0,x:insp-refinedfields-1,x:insp-refinedfields-2} exhibit details that are more similar in structure to the scene, and that are sharper than planes \labelcref{x:insp-kplanes-0,x:insp-kplanes-1,x:insp-kplanes-2}.

\section{Supplementary Results}
\label{x:results}
\begin{figure}[H]
    \centering
    \includegraphics[width=0.85\textwidth]{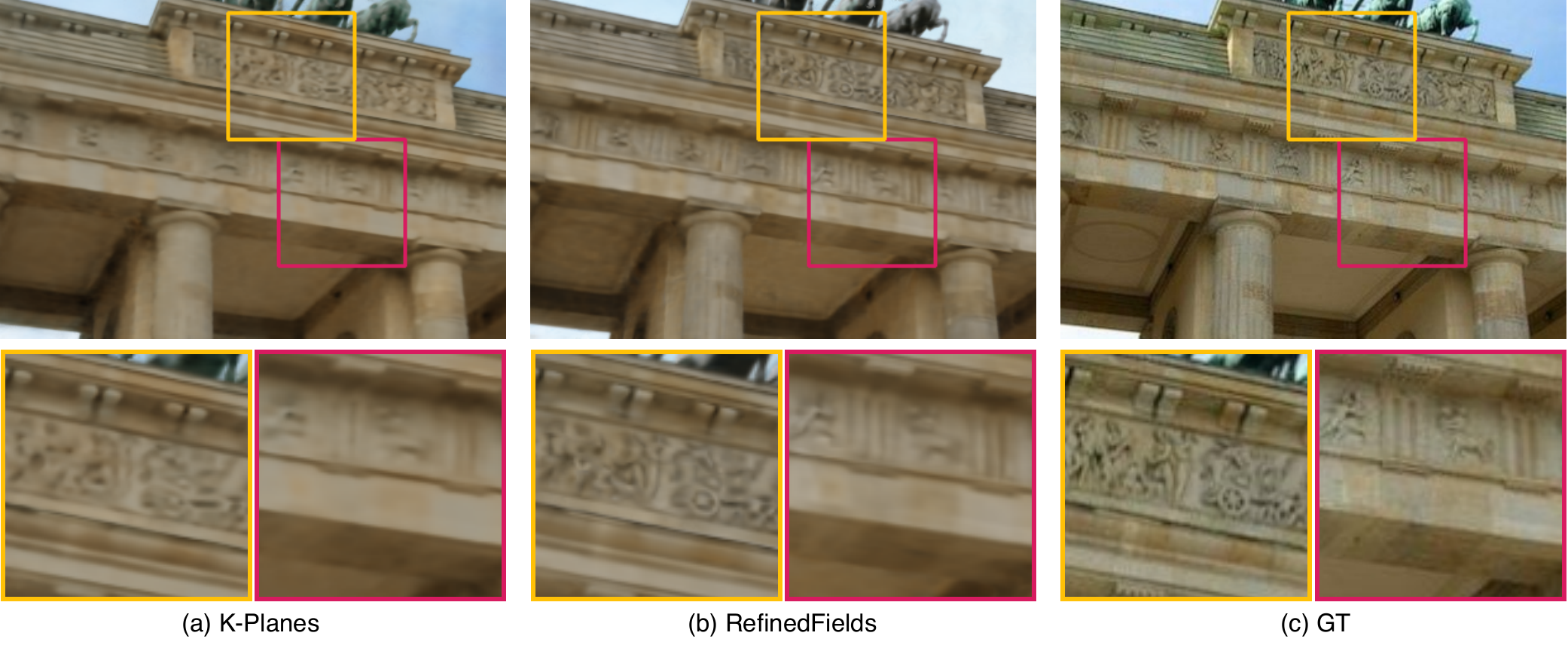}
    \caption{\textbf{Qualitative results.} Results on the \textit{Brandenburg Gate} scene from Phototourism \citep{phototourism}.}
    \label{fig:m-brandenburg}
\end{figure}

\begin{figure}[H]
    \centering
    \includegraphics[width=0.85\textwidth]{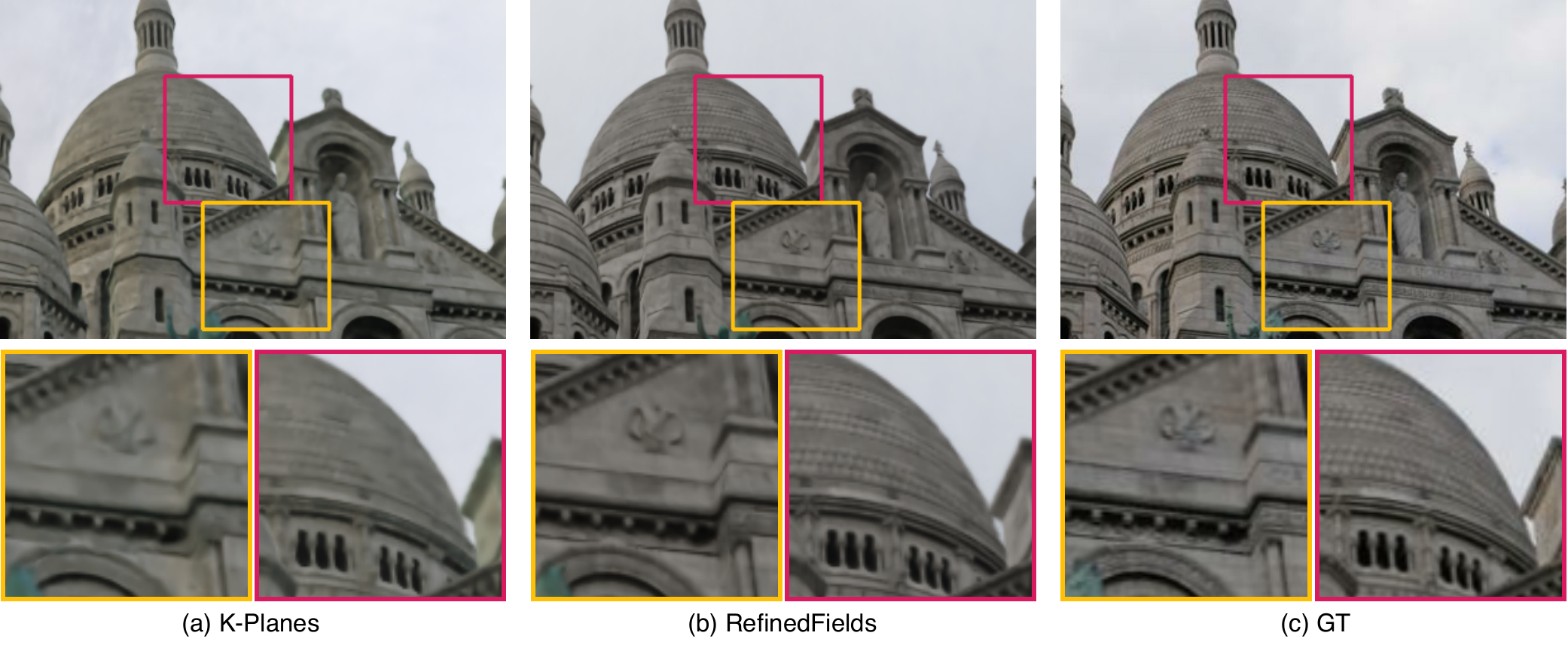}
    \caption{\textbf{Qualitative results.} Results on the \textit{Sacré Coeur} scene from Phototourism \citep{phototourism}.}
    \label{fig:m-sacre}
\end{figure}

\begin{figure}[H]
    \centering
    \includegraphics[width=0.85\textwidth]{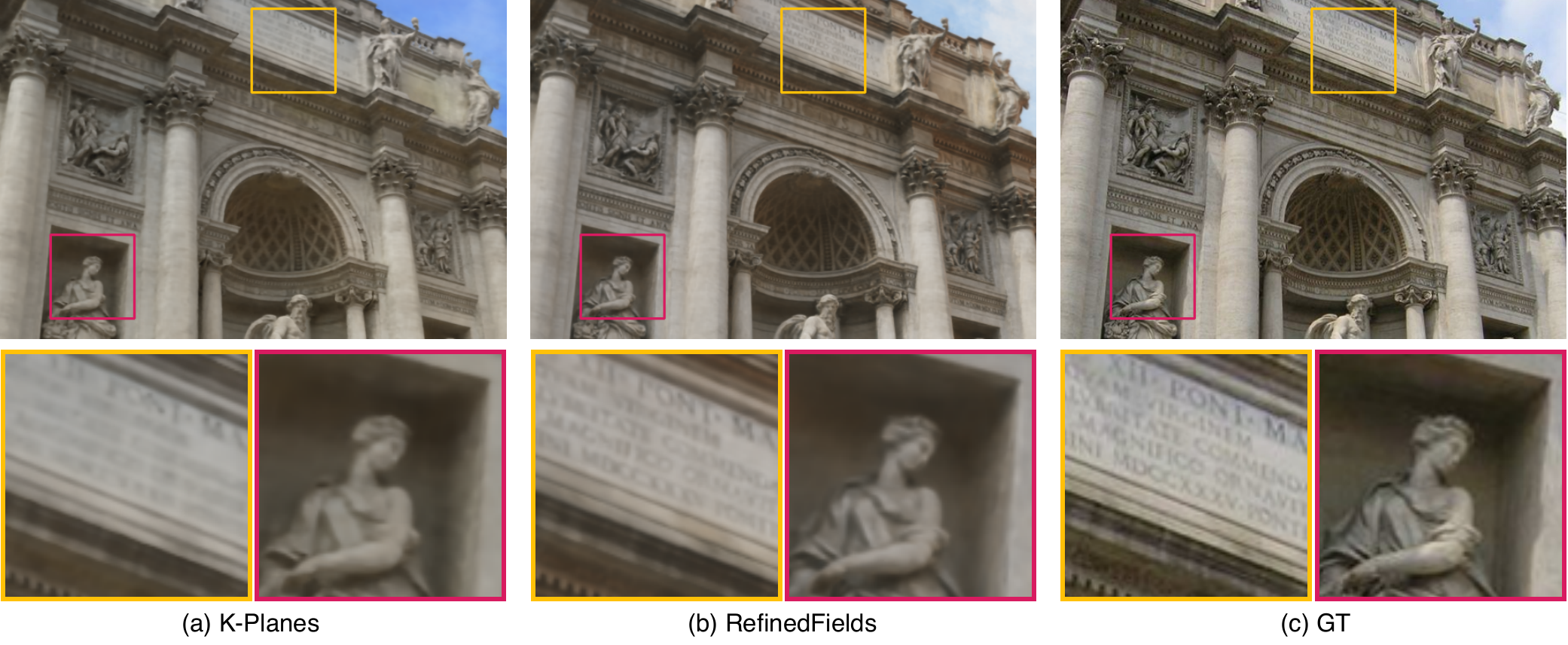}
    \caption{\textbf{Qualitative results.} Results on the \textit{Trevi Fountain} scene from Phototourism \citep{phototourism}.}
    \label{fig:m-trevi}
\end{figure}

\begin{figure}[H]
    \centering
    \includegraphics[width=\textwidth]{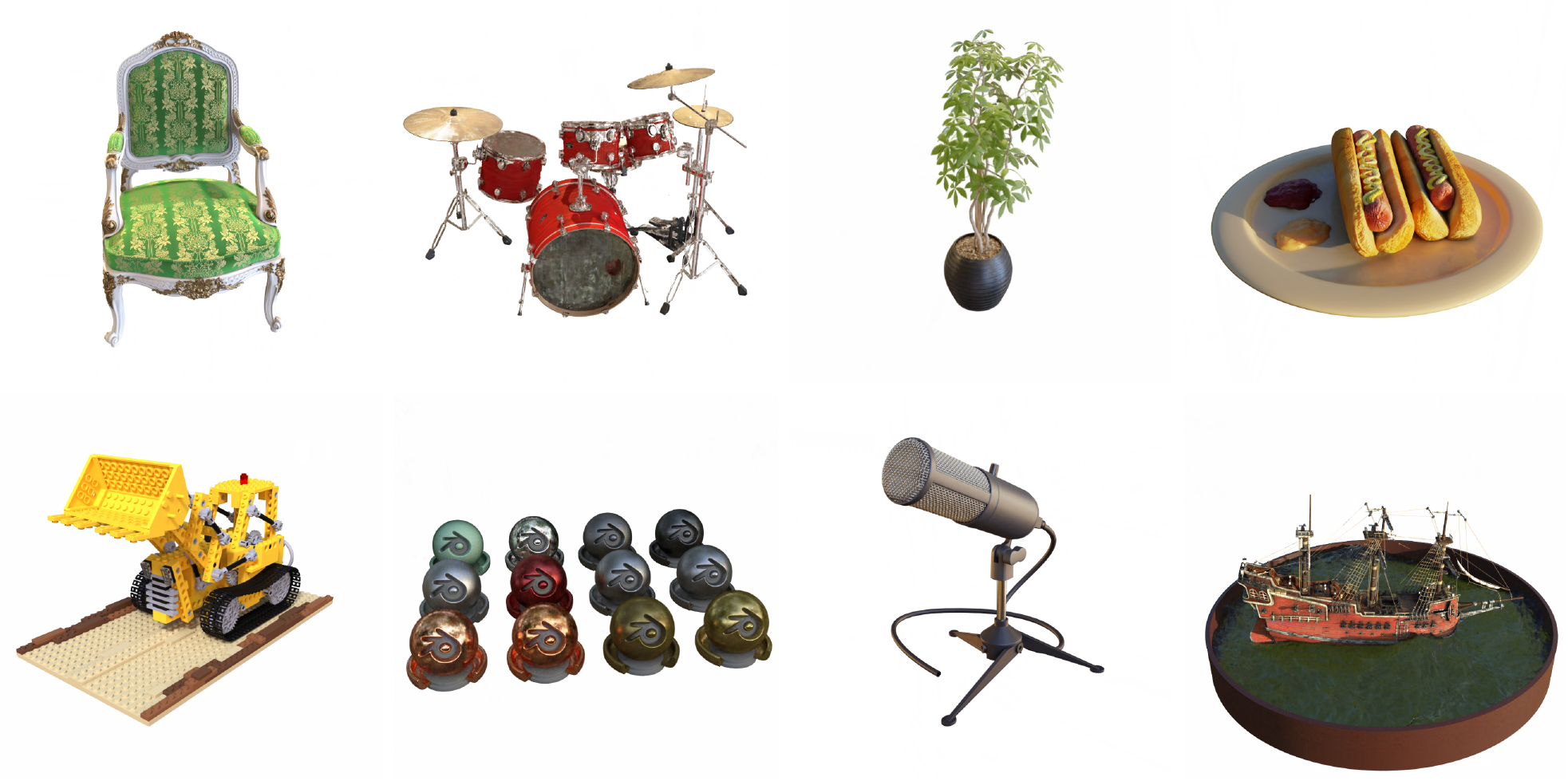}
    \caption{\textbf{Qualitative results.} RefinedFields results on the NeRF Synthetic scenes.}
    \label{fig:m-eve-synth}
\end{figure}

\begin{figure}[H]
    \centering
    \includegraphics[width=\textwidth]{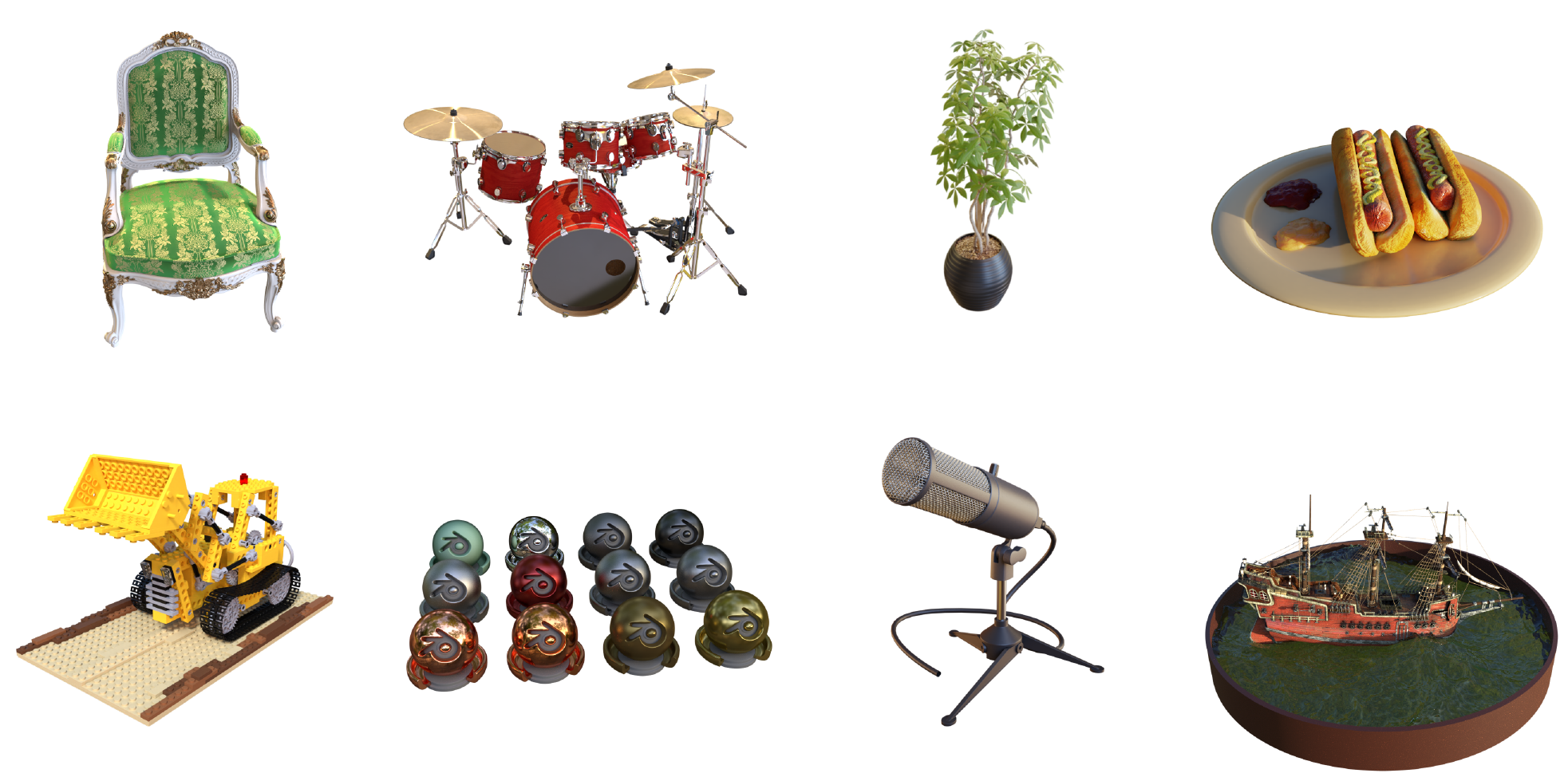}
    \caption{\textbf{Ground Truth Renderings.} Ground truth images from the NeRF Synthetic dataset.}
    \label{fig:m-gt-synth}
\end{figure}

We present below additional qualitative results for in-the-wild scenes (\cref{fig:m-brandenburg,fig:m-sacre,fig:m-trevi}) and synthetic scenes (\cref{fig:m-eve-synth,fig:m-gt-synth}).
\vspace*{\fill}

\newpage
\vspace*{\fill}

\vspace*{\fill}

\newpage
\vspace*{\fill}
\vspace*{\fill}

\newpage
\vspace*{\fill}
\vspace*{\fill}

\newpage
\vspace*{\fill}
\vspace*{\fill}

\newpage
\vspace*{\fill}
\vspace*{\fill}

\newpage
\vspace*{\fill}
\begin{table}[H]
\centering
\footnotesize
\caption{\textbf{Hyperparameters.} A summary of the hyperparameters used to train our model. Appearance optimizations only apply for in-the-wild training.\\
$^{*}$Note that this parameter is taken differently from \citet{kplanes}, as we only work with single-scale planes. We consider the highest plane resolution from the multi-scale approach taken by \citet{kplanes}.}
\vskip 0.15in
{
\begin{tabular}{l c} \toprule
    Hyperparameter & Value \\ \midrule
    Epochs ($N_\mathrm{epochs}$) & \makecell{200 (synthetic) \\ 20 (\emph{Sacré Coeur}) \\ 20 (\emph{Brandenburg Gate}) \\ 10 (\emph{Trevi Fountain}) } \\
    Fitting iterations ($N_1$) & 30000 \\
    Refining iterations ($N_2$) & 3000 \\
    Batch size & 4096 \\
    Optimizer & Adam \\
    Scheduler & Warmup Cosine \\
    K-Planes Learning Rate & 0.01 \\
    LoRA Learning rate & 0.0001 \\
    LoRA rank ($r$) & 4 \\ \midrule
    SD latent resolution & 64 \\
    SD channel dimension & 4 \\
    SD prompt & `` '' \\ \midrule
    Number of planes & 3 \\
    K-Planes resolution$^{*}$ & 512 \\
    K-Planes channel dimension & 32 \\ \midrule
    Epochs Appearance Optimization & 10 \\
    Appearance embeddings dimension & 32 \\
    Appearance learning rate & \makecell{0.1 (\emph{Sacré Coeur}) \\ 0.1 (\emph{Trevi Fountain}) \\ 0.001 (\emph{Brandenburg Gate})} \\
    Appearance batch size & 512 \\ \bottomrule
\end{tabular}
}
\label{table:hyp}
\end{table}
\vspace*{\fill}

\end{document}